%% file: main.tex
\documentclass[times, review, 10pt]{elsarticle}

\usepackage{amssymb}
\usepackage{amsmath}

\usepackage{amsmath,amsfonts}
\usepackage{algorithmic}
\usepackage{array}
\usepackage{textcomp}
\usepackage{stfloats}
\usepackage{url}
\usepackage{verbatim}
\usepackage{graphicx}

\DeclareMathOperator*{\argmin}{argmin}
\usepackage[T1]{fontenc}
\usepackage{hyperref}
\usepackage{chngcntr}
\counterwithout{figure}{section}
\counterwithout{figure}{subsection}
\usepackage{subcaption}
\usepackage{float}
\usepackage[utf8]{inputenc}
\usepackage[ruled,vlined]{algorithm2e}
\usepackage{amsthm}
\usepackage{enumerate}
\usepackage{booktabs}
\usepackage{multirow}

\journal{Preprint}

\begin{document}

\begin{frontmatter}



\title{Msmsfnet: a multi-stream and multi-scale fusion net for edge detection}


\author{Chenguang Liu\fnref{label1}}
\ead{chenguang.liu.light@outlook.com}
\author{Chisheng Wang\corref{cor1}\fnref{label1}}
\ead{wangchisheng@szu.edu.cn}
\author{Feifei Dong\fnref{label1}}
\author{Xiayang Xiao\fnref{label2}}
\author{Xin Su\fnref{label3}}
\author{Chuanhua Zhu\fnref{label1}}
\author{Dejin Zhang\fnref{label1}}
\author{Qingquan Li\fnref{label1}}

\fntext[label1]{School of Architecture and Urban Planning, Shenzhen University, Shenzhen, China; Ministry of Natural Resources (MNR) Key Laboratory for Geo-Environmental Monitoring of Great Bay Area, Shenzhen, China; Guangdong Key Laboratory of Urban Informatics, Shenzhen, China; Guangdong Laboratory of Artificial Intelligence and Digital Economy (SZ), Shenzhen, China.}
\fntext[label2]{China Mobile Internet Company Limited, Guangzhou, China; Key Laboratory for Information Science of Electromagnetic Waves (Ministry of Education), School of Information Science and Technology, Fudan University, Shanghai, China.}
\fntext[label3]{School of Remote Sensing and Information Engineering, Wuhan University, Wuhan, China.}
\cortext[cor1]{Chisheng Wang is the corresponding author.}

\begin{abstract}
Edge detection is a long-standing problem in computer vision. 
Despite the efficiency of existing algorithms, their performance, however, rely heavily on the pre-trained weights of the backbone network on the ImageNet dataset. 
The use of pre-trained weights in previous methods significantly increases the difficulty to design new models for edge detection without relying on existing well-trained ImageNet models,
as pre-training the model on the ImageNet dataset is expensive and becomes compulsory to ensure the fairness of comparison. 
Besides, the pre-training and fine-tuning strategy is not always useful and sometimes even inaccessible. 
For instance, the pre-trained weights on the ImageNet dataset are unlikely to be helpful for edge detection in Synthetic Aperture Radar (SAR) images due to strong differences in the statistics between optical images and SAR images. 
Moreover, no dataset has comparable size to the ImageNet dataset for SAR image processing.
In this work, we study the performance achievable by state-of-the-art deep learning based edge detectors in publicly available datasets when they are trained from scratch, and devise a new network architecture, the multi-stream and multi-scale fusion net (msmsfnet), for edge detection. 
We show in our experiments that by training all models from scratch, our model outperforms state-of-the-art edge detectors in three publicly available datasets.
We also demonstrate the efficiency of our model for edge detection in SAR images, where no useful pre-trained weight is available.
Finally, We show that our model is able to achieve competitive performance on the BSDS500 dataset when the pre-trained weights are used.
\end{abstract}




\end{frontmatter}



\input{introduction}

\input{related_work}
\input{msmsfnet}

\input{experiments}

\input{conclusions}

\section*{Fundings}
This work is supported in part by the National Natural Science Foundation of China (Grant No. 42374018 and Grant No. 42304012), 
in part by the Guangdong Basic and Applied Basic Research Foundation (Grant No. 2025B1515020092 and Grant No. 2022A1515110730), 
in part by the Shenzhen Science and Technology Program (Grant No. KCXFZ20240903093000002 and Grant No. JCYJ20220531101409021).
and in part by the National Natural Science Foundation of China (Grant No. 62371348).


\bibliographystyle{elsarticle-num} 
\bibliography{all_ref}



\end{document}

%% file: introduction.tex
\section{Introduction}
\label{introduction}
Edge detection in both natural images~\cite{sobel, canny, statisticaledge, pb, gpb, structurededge} and remote sensing images~\cite{ROA88, ROEWA98, grhed, lsdsar, csw} has been studied for several decades due to the importance of edge features in image analysis and the use of edges in many applications such as line segment detection~\cite{hough72, hough87, poe}, image segmentation~\cite{gpb, GrabCutIF}, optical flow estimation~\cite{epicflow} and object detection~\cite{gacs, mlrc}. 
Despite the large amount of research dedicated to the problem of edge detection, state-of-the-art edge detection algorithms mainly rely on deep learning. 
However, their performance heavily relies on the pre-trained weights of the backbone network on the ImageNet dataset~\cite{imagenet}. 
Apparently their performance will drop if the pre-trained weights are not used, as much smaller amount of data is used for training. 
As is known to all, the following four aspects are of crucial importance for the performance of deep learning models: the training data, network architecture, loss function and the training strategy (optimizers, hyperparameter settings). 
The use of pre-trained weights on the ImageNet dataset in most existing edge detectors increases the difficulty to design new network architectures for edge detection without relying on existing well-trained ImageNet models, 
as it is quite expensive to train a model on the ImageNet dataset (the ImageNet-1k dataset contains around 1.28 million images and the ImageNet-21k dataset contains around 14 million images), and thus not feasible for many researchers due to limited computation resources. 

On the other hand, the pre-trained weights of existing ImageNet models (such as VGG net ~\cite{vgg} and ResNet ~\cite{resnet}) are unlikely to improve the edge detection accuracy if the aim is to train the model for edge detection in a different kind of images, Synthetic Aperture Radar (SAR) images for example, because of strong differences in the statistics between SAR and natural images ~\cite{grhed}. 
Natural images are usually contaminated by weak, additive Gaussian noise. 
In contrast, SAR images are corrupted by strong multiplicative noise. 
In a real SAR image, the intensity probability density function (pdf) of a homogeneous area follows a Gamma distribution under the hypothesis of fully developed speckle ~\cite{goodman75}.
Another strong distinction between a natural image and a real SAR image is the range of pixel values.
In a natural image, the maximum pixel value is often 255, while in a real SAR image, the maximum pixel value could be 10000 or much larger.
These strong distinctions determine that the pre-trained weights of the backbone networks on the ImageNet dataset are unlikely to help to improve the performance of existing deep learning methods for edge detection in real SAR images.
Furthermore, no dataset in the field of SAR image processing has comparable size to the ImageNet dataset. 
The pre-training (in a very large dataset) and then fine-tuning (in a small dataset) strategy commonly used in the field of computer vision is thus not suitable for SAR image processing.
Hence, it's indeed meaningful to study the performance achievable by deep learning based edge detectors without using pre-trained weights.

To facilitate the design of new network architectures for edge detection in future work, we set new benchmarks to evaluate edge detection algorithms in publicly available datasets. 
We train state-of-the-art deep learning based edge detectors in three publicly available datasets for edge detection~\cite{bsds500, nyudv2, dexined} and compute the optimal performance that can be achieved by them when training from scratch. 
We then propose a new network architecture for edge detection, the multi-stream and multi-scale fusion net (msmsfnet), and show that it outperforms these edge detectors when training from scratch in the three datasets. 
The efficiency of the proposed method is then demonstrated for edge detection in SAR images, in which case no useful pre-trained weight is available.
Finally, we show in our experiments that with the help of pre-trained weights from the ImageNet dataset, the performance of our model on the BSDS500 ~\cite{bsds500} dataset improves further.
This pre-training and fine-tuning experiment provides additional evidence to support our arguments that: 1) the performance of deep learning models rely heavily on the amount of training data; 2) the comparison of the performance of different algorithms should be done under the same experimental settings (i.e., the same amount of data should be used for training) and would be unfair otherwise.

The contributions of this work are summarized as follows:
\begin{itemize}
\item We set new benchmarks to evaluate edge detection algorithms in publicly available datasets, which facilitates the design of new network architectures for edge detection.
  \item We propose a new network architecture for edge detection, which outperforms state-of-the-art edge detectors in three publicly available datasets when all models are trained from scratch, and the proposed method is able to achieve competitive performance when the pre-trained weights are used.
  \item The efficiency of the proposed method is further demonstrated for edge detection in SAR images, where no useful pre-trained weight is available for existing methods. This serves as an important evidence showing that it's indeed meaningful to study the performance that can be achieved by deep learning based edge detectors when they are trained from scratch.
  \end{itemize}

The rest of the paper is organized as follows: 
Section~\ref{related_work} gives a brief review of edge detection algorithms for both natural images and SAR images;
Section~\ref{sec::msmsfnet} provides the details about the proposed network architecture; 
in Section~\ref{experiments}, we demonstrate the efficiency of the proposed method by comparing it with state-of-the-art deep learning based edge detectors in both natural images and SAR images;
Conclusions are finally given in Section~\ref{conclusions}.

%% file: related_work.tex
\section{Related work}
\label{related_work}
\subsection{Edge detection in natural images}
Since state-of-the-art methods for edge detection in natural images are mainly based on deep learning, we only give a brief review of the deep learning based edge detection algorithms.

In the early stage, the CNN-based edge detectors learned to detect edges from image patches~\cite{deepedge, deepcontour}, until the proposal of the Holistically-nested edge detection (HED)~\cite{hed_ijcv}, which used a fully convolutional design and could thus be trained end-to-end and performed image-to-image prediction. 
Besides, the deep supervision technique~\cite{dsn} used in HED enabled the network to learn and fuse multi-scale and multi-level features, which were important for extracting edges from various scales. 
Later on, various approaches were proposed following the fashion of HED, but with more complicated techniques to enhance the multi-scale feature learning and fusion capability of the proposed methods. 
CEDN~\cite{cedn} trained a fully convolutional encoder-decoder network (the encoder was initialized using the VGG-16 network~\cite{vgg} and followed by a light-weight decoder network), in order to fully exploit the multi-scale representations learned by the pre-trained VGG-16 net for object contour detection. 
Starting from the pre-trained VGG net~\cite{vgg} and ResNet~\cite{resnet}, COB~\cite{cob} showed that the boundaries could be detected more accurately by learning jointly the contour strength and contour orientations in multi-scale. 
RCF~\cite{rcf_pami} proposed to combine the features extracted from all convolutional layers instead of using only the last layer of each stage as in HED, so as to fully exploit the multi-scale and multi-level information learned by all convolutional layers. 
By using more training data, the multi-scale testing technique and the improved network architecture, RCF achieved better performance than HED in the publicly available datasets for edge detection. 
CED~\cite{ced} proposed a novel refinement network architecture, which increased the resolution of feature maps progressively with sub-pixel convolution, in order to improve the crispness of edges detected by HED. 
DeepBoundaries~\cite{deep_boundaries} extended HED with a multi-resolution architecture, and improved the edge detection accuracy of HED by using more training data and a well designed loss function. 
The performance of DeepBoundaries was further improved by combining the proposed multi-resolution architecture with the Normalized Cuts technique~\cite{ncis}. 
Similar to CED, the RefineContourNet (RCN)~\cite{rcn} was also designed following the idea of RefineNet~\cite{refinenet}, which used a multi-path refinement architecture, for more effective utilization of the high-level abstraction ability of the ResNet. 
BDCN~\cite{bdcn} adopted a bi-directional cascade network structure, which supervised individual layer with edges at its specific scale, to extract edges at different scales. 
A Scale Enhancement Module (SEM) was also proposed in BDCN in order to enrich the multi-scale representations of each layer by using dilated convolutions. 
CATS~\cite{cats} proposed to improve the localization accuracy of existing deep edge detectors with a novel tracing loss and a context-aware fusion block, tackling the problem of feature mixing and side mixing of existing methods.
Apart from these methods, DexiNed~\cite{dexined} proposed a new network architecture for edge detection and created a new benchmark dataset (BIPEDv2). 
It showed that by training all models from scratch in the BIPEDv2 dataset, DexiNed outperformed all existing CNN-based edge detectors. 
Different from previous methods, DexiNed was not built on the basis of VGG net or the ResNet. 
Therefore, no pre-trained weights on the ImageNet dataset is available for DexiNed, which could be an important reason why it does not provide quantitative comparisons on previous boundary detection datasets.
UAED ~\cite{uaed} proposed to improve the edge detection accuracy by measuring the uncertainty and investigating the subjectivity and ambiguity among different annotations of the edge ground truth.
MUGE ~\cite{muge} improved further the performance of UAED by estimating the edge granularity of individual annotation, which could be used to guide the generation of diversified edge maps at different controllable granularities.
Despite the efficiency of UAED and MUGE, the strategies proposed by them are applicable only when multiple annotations are available for each training image.
EDTER ~\cite{edter} proposed a transformer-based edge detector to fully exploit the long-range dependencies captured by vision transformer ~\cite{vit}, and achieved competitive performance in several benchmark datasets for edge detection.
However, the performance achieved by EDTER heavily rely on the pre-trained weights of the backbone network ~\cite{vit} on the ImageNet-22k dataset, namely the ImageNet-21k dataset plus the ImageNet-1k dataset.
Much more data is actually used to train the model than previous methods.
As what we will show in the experimental section (Subsection ~\ref{pre_fine}), the performance of EDTER is far below than that of CNN-based edge detectors when training from scratch.

\subsection{Edge detection in SAR images}
Edge detection in SAR images has been studied for several decades due to the importance of edge features in image analysis, line segment detection~\cite{poe}, road extraction~\cite{florence98} and many other applications~\cite{csw}. 
Although many edge detectors are efficient for optical images~\cite{sobel, canny, gpb, structurededge}, they are, however, not suitable for edge detection in SAR images, mainly because of the strong differences in the statistics of optical and SAR images~\cite{ROA88}. 
It has been carefully studied and shown in ~\cite{ROA88} that the usual difference based gradient operators, which are commonly used in edge detectors for optical images, do not have a constant false alarm rate (CFAR) for SAR images, while the Ratio of Average (ROA)~\cite{ROA88} is more suitable to perform edge detection in SAR images thanks to its CFAR property. 
Later on, the exponentially weighted average was proved to be optimal to estimate the local mean value of a homogeneous area under the hypothesis of a stochastic multi-edge model, and the Ratio of Exponentially Weighted Average (ROEWA) was proposed to perform edge detection in SAR images~\cite{ROEWA98}. 
~\cite{GGS12} and ~\cite{crater16} were still based on the ratio operation, but use different shapes of window functions to estimate the local mean value of homogeneous areas. 
Despite the valuable CFAR property of these ratio-based edge detectors, the edge detection results obtained by them were not satisfying enough because of the strong multiplicative noise in SAR images, especially in the most challenging 1-look situation. 
To improve the edge detection performance in 1-look real SAR images, a deep learning based edge detector GRHED was proposed in ~\cite{grhed}. 
Due to the absence of a real SAR dataset to train deep learning models for edge detection, it proposed a strategy to enable models trained using simulated SAR images to be efficient for real SAR images. 

%% file: msmsfnet.tex
\section{Msmsfnet: a multi-stream and multi-scale fusion net for edge detection}
\label{sec::msmsfnet}
\subsection{Motivation}
Developing deep convolutional neural networks for computer vision tasks has become popular ever since the proposal of Alexnet~\cite{alexnet}, which emphasizes the importance of depth on the performance of CNNs. 
Alexnet consists of five convolutional layers and three fully connected layers, among which kernels of large sizes ($11\times 11$ and $5\times 5$) are used in the first two convolutional layers. 
Afterwards, the very deep convolutional networks (VGG net) was proposed in ~\cite{vgg}, which utilized small convolutional filters ($3\times 3$) in all convolutional layers. 
As described in ~\cite{vgg}, a sequence of $3\times 3$ convolutional filters has the same effective receptive field as convolutional filters of larger size, while it has fewer parameters to learn when the number of input and output channels is equal. 
The VGG net~\cite{vgg} showed that the accuracy of the network on the large scale image recognition tasks can be significantly boosted by pushing the depth of the network to 19 weight layers. 
Parallel to the work of VGG net, GoogLeNet~\cite{googlenet} (also known as inceptionv1) achieved similarly high performance to the VGG net but with much more complicated network architecture. 
GoogLeNet pushed the depth of the network to 22 weight layers through the excessive use of dimension reduction with $1\times 1$ convolutions. 
Dimension reduction using $1\times 1$ convolution was also widely used in subsequent inception network architectures~\cite{inceptionv2, inceptionv3, inceptionv4}, the ResNet and its variants~\cite{resnext, res2net}, which reduces the number of parameters to be learned in each layer through the use of $1\times 1$ convolution and thus makes it possible to increase the depth of the network significantly. 
Besides, the number of parameters in the network can also be reduced by factorizing large convolutional layers into small ones or by factorizing them into spatial asymmetric convolutions~\cite{inceptionv3}.

As demonstrated in previous work~\cite{alexnet, vgg, googlenet, resnet, inceptionv2, inceptionv3, inceptionv4}, the performance of CNN models can be boosted significantly by increasing the depth of the network, while a crucial factor to develop very deep CNNs is to reduce the number of parameters in each layer. 
In this work, we follow the strategy used in~\cite{inceptionv3} and factorize usual convolutional filters into spatial asymmetric convolutions so as to reduce the number of parameters to be learned in each layer. 
More specifically, a $n\times n$ convolutional layer can be factorized into a $1\times n$ convolutional layer followed by a $n\times 1$ convolutional layer (or the reverse). 
This factorization reduces significantly the number of parameters to be learned while maintaining the size of the receptive field. 
For instance, factorizing a $3\times 3$ convolution into a $1\times 3$ convolution followed by a $3\times 1$ convolution contributes to a $33$ percent reduction in the number of parameters for the same number of input/output filters (assuming that the number of input and output filters is equal). 
Much more computational cost can be saved for the factorization of filters with larger kernel sizes. 
We utilize spatial asymmetric convolutions excessively  in our network and push the depth of our proposed model to 74 weight layers.

\begin{figure}[H]
  \centering
  \begin{tabular}{c}
    \includegraphics[width=0.48\textwidth]{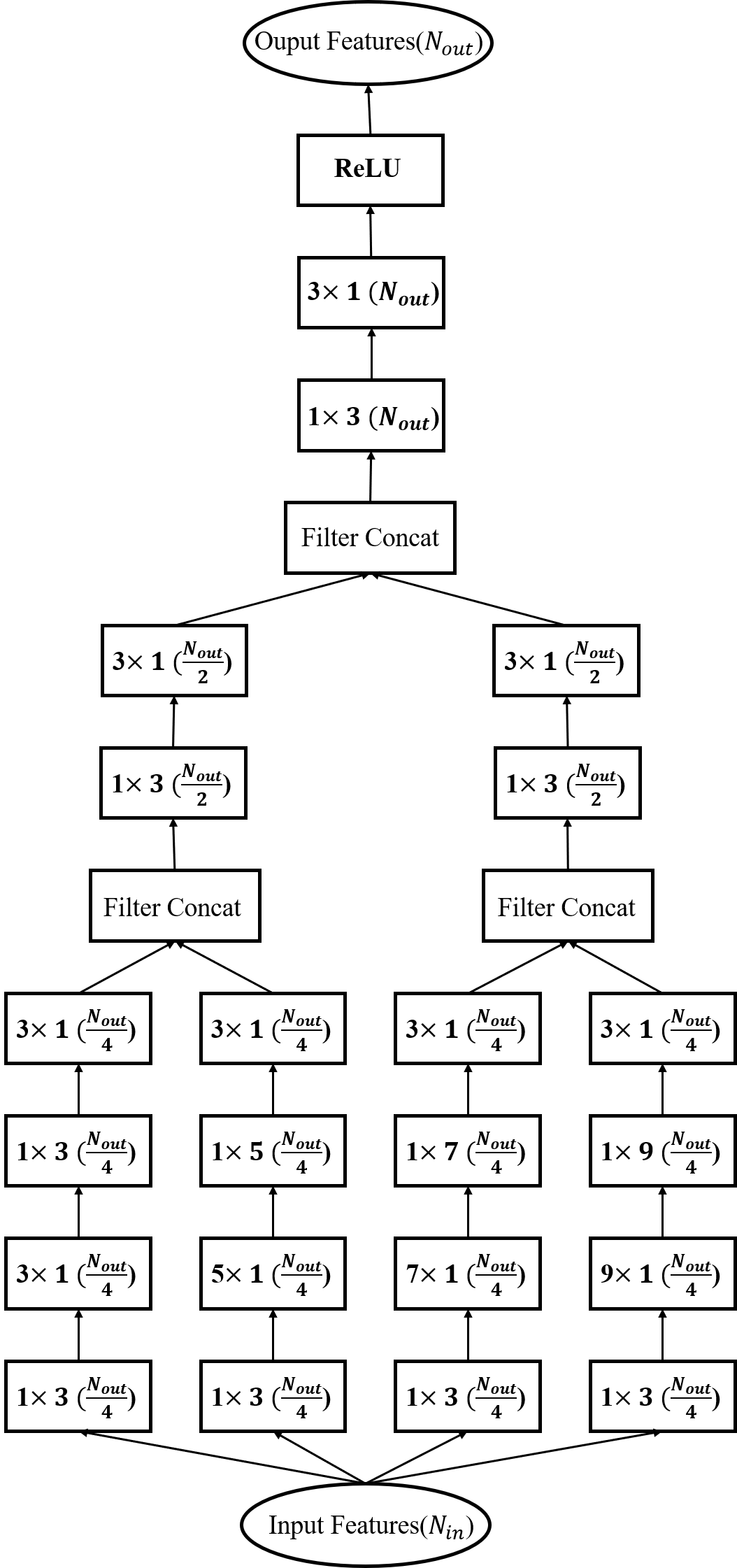}
  \end{tabular}
  \caption{The proposed msmsfblock. The expression in the parenthesis indicates the number of output filters of the layer.}
  \label{msmsfblock}
  \end{figure}

\subsection{Network architecture}
For the task of edge detection, the multi-scale representation capability of models is of crucial importance. 
Existing CNN-based edge detectors constantly boost the performance of edge detection by enriching the multi-scale representations learned by the backbone networks in various ways. 
Although CNN models are naturally multi-scale/multi-level feature extractors because of the downsampled strided layers as well as the increasingly larger receptive field, we can still improve the multi-scale representation capability of CNNs by exploiting network architectures of higher quality.

To enrich the multi-scale representations, we devise a multi-stream and multi-scale fusion block (msmsfblock), and construct our msmsfnet by simply stacking our proposed msmsfblock. 
As shown in Figure~\ref{msmsfblock} which displays the configurations of our msmsfblock, the input feature maps are fed into four parallel branches, extracting features at different scales due to the difference in the size of the receptive field. 
The extracted features are then fused by a two-step fusion stage, where the first step fuses features of closer scales, and the second step produces the final output of the msmsfblock. 
In our msmsfblock, we utilize spatial asymmetric convolutions excessively ($1\times n$ and $n\times 1$ convolutions), and only the final $3\times 1$ convolutional filter is followed by a ReLU function~\cite{alexnet}. 
Batch normalization~\cite{inceptionv2} is not used in our msmsfblock.

\begin{figure}[h]
  \centering
  \begin{tabular}{c}
  \includegraphics[width=0.48\textwidth]{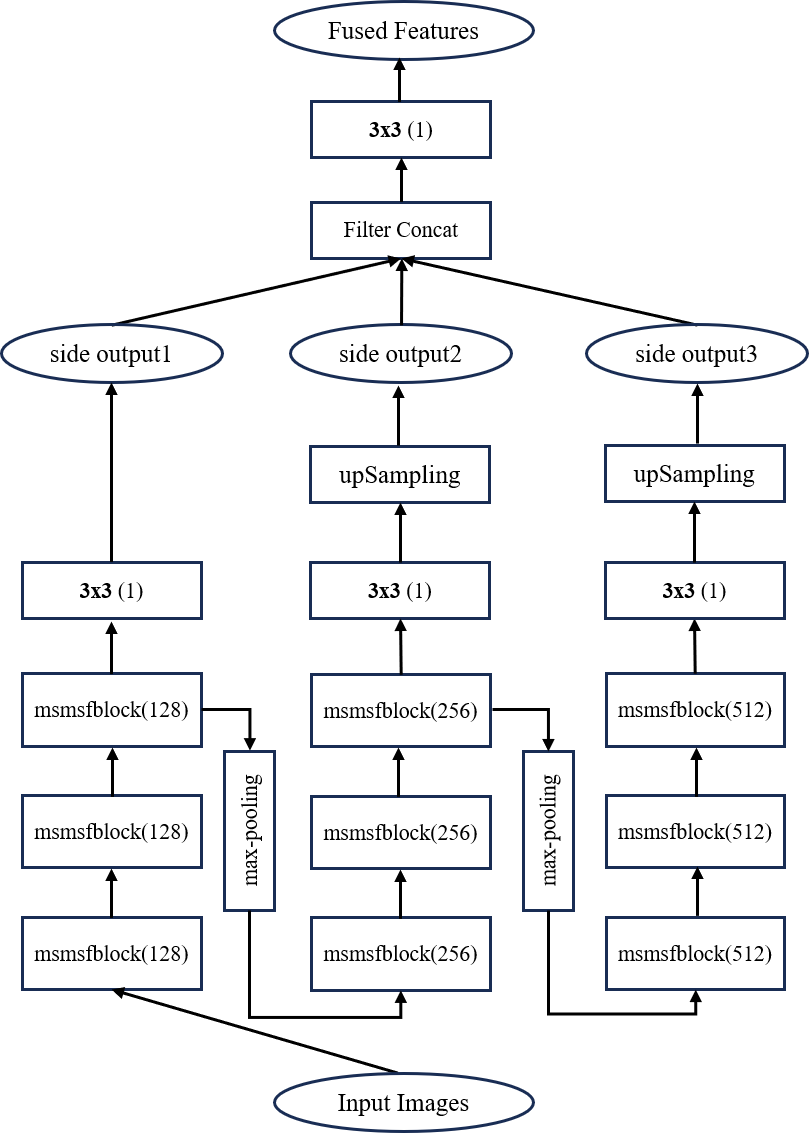} \\
  \end{tabular}
  \caption{Network architecture of the proposed msmsfnet. The number in the parenthesis indicates the number of output filters of the layer.}
  \label{msmsfnet}
  \end{figure}

Our proposed msmsfnet is built by stacking the msmsfblock, with spatial max-pooling layers (kernel size $3\times 3$, stride 2) injected in-between. 
Similar to HED~\cite{hed_ijcv}, we adopt the deep supervision technique~\cite{dsn}. 
We produce side outputs by adding side layers to the msmsfblocks before max-pooling layers and the final msmsfblock. 
In the side layers, a $3\times 3$ convolution is used to reduce the channel dimension of the feature maps to 1. 
If the feature maps have reduced resolution due to the downsampling operation done by the max-pooling layer, a deconvolutional layer which is fixed to perform bilinear interpolation is used to upsample the feature maps to the resolution of the input image. 
The final output of our msmsfnet are then obtained by fusing these side outputs with a $3\times 3$ convolutional layer. 
All side outputs and the fused output are supervised by the edge ground truth with a weighted cross entropy loss as defined in ~\cite{rcf_pami}. 
Details about our proposed msmsfnet are displayed in figure~\ref{msmsfnet}.

\subsection{Formulation}
\subsubsection{Training phase}
Given an image $u$ in the training dataset, notating $\mathbb{G}$ as its associated edge ground truth, where $u=\{ u_j, j=1, \cdots, |u|\}$ denotes the input image and $\mathbb{G}=\{ \mathbb{G}_j, j=1, \cdots, |\mathbb{G}|\}$ denotes the edge ground truth. 
Notating $\mathbb{W}$ the collection of parameters in all standard network layer (excluding those corresponding to side layers and the final fusion layer), and notating $\mathbb{W}_{side}=(\mathbb{W}_{side}^{(1)}, \mathbb{W}_{side}^{(2)}, \mathbb{W}_{side}^{(3)})$ the collection of parameters in those side layers, the objective function for the side layers is defined as:
\begin{eqnarray}
\mathcal{L}_{side}(\mathbb{W},\mathbb{W}_{side})=\sum_{m=1}^{3}\beta_m\ell_{side}^{(m)}(\mathbb{W},\mathbb{W}_{side}^{(m)}),
\label{objective_side}
\end{eqnarray} 

where $\ell_{side}^{(m)}$ denotes the loss corresponding to the $m$-th side output.

In order to balance the loss between positive and negative classes, a class-balanced cross-entropy loss function as defined in ~\cite{rcf_pami} is used in formula~\eqref{objective_side}:

\begin{eqnarray}
  \ell_{side}^{(m)}(\mathbb{W},\mathbb{W}_{side}^{(m)}) =  -\lambda\sum_{j\in \mathbb{G}_{+}}\log\mathbb{P}(\mathbb{G}_j=1 |u; \mathbb{W}, \mathbb{W}_{side}^{(m)}) &\nonumber \\ 
   -1.1\times (1-\lambda)\sum_{j\in \mathbb{G}_{-}}\log\mathbb{P}(\mathbb{G}_j=0 | u; \mathbb{W},\mathbb{W}_{side}^{(m)}),&  
\end{eqnarray}  

where $\lambda=\frac{|\mathbb{G}_{-}|}{|\mathbb{G}|}$, and $1-\lambda=\frac{|\mathbb{G}_{+}|}{|\mathbb{G}|}$, $|\mathbb{G}_{-}|$ and $|\mathbb{G}_{+}|$ represent the non-edge and edge label sets. 
$\mathbb{P}(\mathbb{G}_j=1 |u; \mathbb{W},\mathbb{W}_{side}^{(m)})=\sigma(a_j^{(m)})\in [0,1]$ with $\sigma(\cdot)$ representing the sigmoid function and $a_j^{(m)}$ representing the activation value at pixel $j$. 
The edge probability map produced by each side layer is thus defined as:
\begin{eqnarray}
\hat{\mathbb{G}}_{side}^{(m)}=\sigma(\hat{A}_{side}^{(m)}),
\end{eqnarray}

where $\hat{A}_{side}^{(m)}\equiv\{a_j^{(m)}, j=1,2,\cdots, |\mathbb{G}|$ represents the activations of the output of the side layer.

The final fused output of the model is obtained by fusing these side outputs with a $3\times 3$ convolutional layer. 
The same loss function as the side layers is used to compute the loss between the edge ground truth and the final fused output.

Denoting $\mathbb{W}_{fuse}$ the collection of parameters corresponding to the fusion layer and $\mathcal{L}_{fuse}(\cdot)$ the loss function corresponding to the fused output, the objective function that needs to be minimized during training is thus defined as:
\begin{eqnarray}
  (\mathbb{W},\mathbb{W}_{side},\mathbb{W}_{fuse})^\star  = & \nonumber \\
  \argmin (\mathcal{L}_{side}(\mathbb{W},\mathbb{W}_{side})+\mathcal{L}_{fuse}(\mathbb{W},\mathbb{W}_{side},\mathbb{W}_{fuse})).  &
\end{eqnarray}

\subsubsection{Testing phase}
Given a testing image $u$, four edge probability maps (three side outputs and one fused output) are obtained from our msmsfnet:
\begin{eqnarray}
  (\hat{\mathbb{G}}_{fuse}, \hat{\mathbb{G}}_{side}^{(1)}, \cdots , \hat{\mathbb{G}}_{side}^{(3)})&=& \nonumber \\
  \mathrm{CNN}(u,(\mathbb{W},\mathbb{W}_{side}, \mathbb{W}_{fuse})^\star). &&
\end{eqnarray}

The fused output is chosen as the final output of our network.

%% file: experiments.tex
\section{Experiments}
\label{experiments}
In this section, we demonstrate the efficiency of our proposed msmsfnet by comparing it with several representative CNN-based edge detectors (HED~\cite{hed_ijcv}, RCF~\cite{rcf_pami}, BDCN~\cite{bdcn}, CATS~\cite{cats}, DexiNed~\cite{dexined}) in both natural images (three publicly available datasets for edge detection, the BIPEDv2 dataset~\cite{dexined}, the BSDS500 dataset~\cite{bsds500} and the NYUDv2 dataset~\cite{nyudv2}) and SAR images (a 1-look SAR dataset simulated using optical images and a 1-look real SAR image). 
We show in our experiments that our method achieves better performance than state-of-the-art CNN-based edge detectors in all these experiments when all models are trained from scratch.
We also show in our experiments that our model is able to achieve competitive performance on the BSDS500 dataset when the pre-trained weights from the ImageNet dataset are used.
The proposed method is not compared to UAED ~\cite{uaed} and MUGE ~\cite{muge} for the following reasons:
first, the contributions of UAED ~\cite{uaed} and MUGE ~\cite{muge} lie mainly on the strategies they proposed to fully exploit the potential of datasets providing multiple annotations for each image and not on the network architecture;
second, the strategies they proposed are applicable only when multiple annotations are available for each image and thus not widely useful;
third, only one annotation is provided for each image in the BIPEDv2 ~\cite{dexined} and NYUDv2 ~\cite{nyudv2} datasets;
fourth, since our work focuses on the design of new network architecture for edge detection, the comparison between the performance of different algorithms should be done under the same experimental settings and would be unfair otherwise.
Comparisons with the transformer based edge detector EDTER ~\cite{edter} will be provided in Subsection~\ref{pre_fine}.

\subsection{Implementation details}
The proposed msmsfnet is implemented using the PyTorch library, and trained from scratch using the Adam optimizer. 
The weights of all convolutional layers are randomly initialized using the Xavier uniform distribution and the biases are all initialized to 0. 
Weight decay is set to $10^{-12}$. 
The mini-batch size for training is set to 6. 
The learning rate is initialized to $10^{-4}$, and decreased by 10 after a fixed number of epochs (10 for the BIPEDv2 dataset, 20 for the BSDS500, NYUDv2 and the simulated SAR dataset). 
The maximum number of epochs to train is set to 15 for the BIPEDv2 dataset, and 25 for the BSDS500, NYUDv2 and the simulated SAR dataset. 
The optimal performance obtained by each method during training is reported in the following. 
The implementation details of the pre-training and fine-tuning experiment on the BSDS500 dataset can be found in Subsection~\ref{pre_fine}.

\subsection{Evaluation criteria}
The outputs of all CNN-based edge detectors are edge probability maps instead of binary edge maps. 
Non-maxima suppression, thresholding and edge thinning are still required to obtain binary edge maps. 
Following previous work~\cite{hed_ijcv, rcf_pami, bdcn}, we adopt the non-maxima suppression step of ~\cite{structurededge} to process the outputs of our model. 
Three criteria are used to compare the performance of different methods: the Optimal Dataset Scale (ODS) F1-score, where a global optimal threshold is chosen to binarize the edge probability maps; 
the Optimal Image Scale (OIS) F1-score, where an optimal threshold is chosen for each image to binarize the edge probability maps; 
and the Average Precision (AP). 
Higher values of these criteria indicate better performance obtained by a method. 
Following previous work, precision-recall curves are also ploted to compare the performance of different methods. 
In the following, we will compare the performance of our msmsfnet with state-of-the-art deep learning based edge detectors using these criteria in three publicly available datasets, and a 1-look simulated SAR dataset.

\subsection{BIPEDv2 dataset}
We first compare the performance of our proposed msmsfnet with HED~\cite{hed_ijcv}, RCF~\cite{rcf_pami}, BDCN~\cite{bdcn}, CATS~\cite{cats} and DexiNed~\cite{dexined} in the second version of the Barcelona Images for Perceptual Edge Detection (BIPEDv2) dataset~\cite{dexined}, 
as the performance reported by existing algorithms on it does not involve the use of pre-trained weights on the ImageNet dataset. 

\begin{table}[h]
  \centering
  \caption{Quantitative comparison of different algorithms on the 50 testing images of the BIPEDv2 dataset.}
  \label{f1_biped}
  \setlength{\tabcolsep}{3pt}
  \begin{tabular}{c c c c}
    \toprule
    Methods & ODS F1-score & OIS F1-score & AP \\
    \midrule
    HED & 0.884 & 0.890 & 0.919 \\
    \midrule
    RCF & 0.882 & 0.889 & 0.911 \\
    \midrule
    BDCN & 0.893 & 0.900 & 0.922 \\
    \midrule
    CATS & 0.894 & 0.899 & 0.927 \\
    \midrule
    DexiNed & 0.894 & 0.900 & 0.921 \\
    \midrule
    msmsfnet & 0.897 & 0.901 & 0.936 \\
    \bottomrule
    \end{tabular}
  \end{table}

The BIPEDv2 dataset contains 200 images for training and 50 images for testing. 
All images are outdoor images and of size $1280\times 720$ pixels. 
We use the data augmentation strategy described in ~\cite{dexined} to augment the training dataset, and train all models from scratch using the augmented training dataset. 
Quantitative comparisons of different methods on the 50 testing images of the BIPEDv2 dataset are displayed in table~\ref{f1_biped}. 

\begin{figure}[h]
  \centering
  \begin{tabular}{c}
    \includegraphics[width=0.5\textwidth]{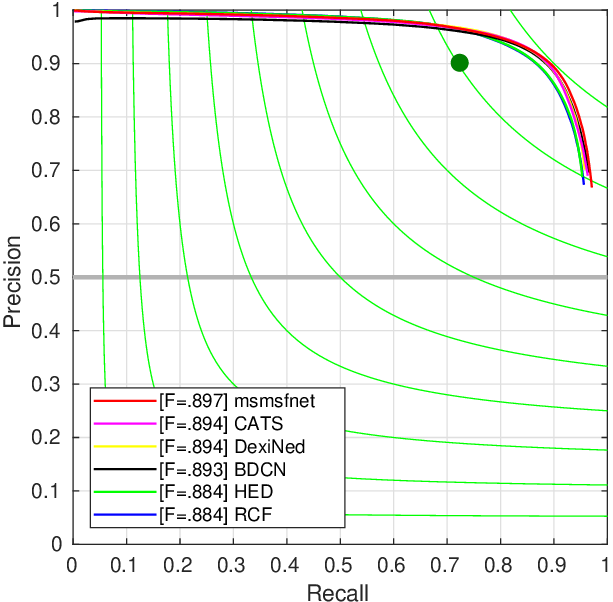}\\
  \end{tabular}
  \caption{The precision-recall curves computed by different methods in the test set of the BIPEDv2 dataset.}
  \label{biped_pr_curve}
\end{figure}

From table~\ref{f1_biped} we can see that the proposed msmsfnet obtains the highest scores in all three criteria. 
Minor improvements over existing methods (0.3 percent for ODS F1-score and 0.1 percent for OIS F1-score) are made by our msmsfnet on the test set of BIPEDv2 dataset. 
The improvements made by our method are relatively larger for the Average Precision (at least 0.9 percent). 
The precision-recall curves computed by different methods in the test set of BIPEDv2 dataset are displayed in figure~\ref{biped_pr_curve}.

As shown in table~\ref{f1_biped} and figure~\ref{biped_pr_curve}, the proposed msmsfnet achieves slightly better performance than existing methods in the BIPEDv2 dataset. 
This is probably because the edge anotations in the dataset are 'low level edges', while the multi-representation capability of a model plays a more important role in higher level tasks such as boundary/contour detection. 
As what we will show in the following, much larger improvements are obtained by the proposed msmsfnet in the boundary/contour detection datasets. 
Details about the differences among edges, boundaries, and contours can be found in ~\cite{dexined} (following previous work~\cite{hed_ijcv, rcf_pami, cob, bdcn, cats}, we do not strictly distinguish between edge and boundary/contour detection and call all these tasks as edge detection).

\begin{figure}[h]
  \centering
  \begin{tabular}{cc}
    \includegraphics[width=0.3\textwidth]{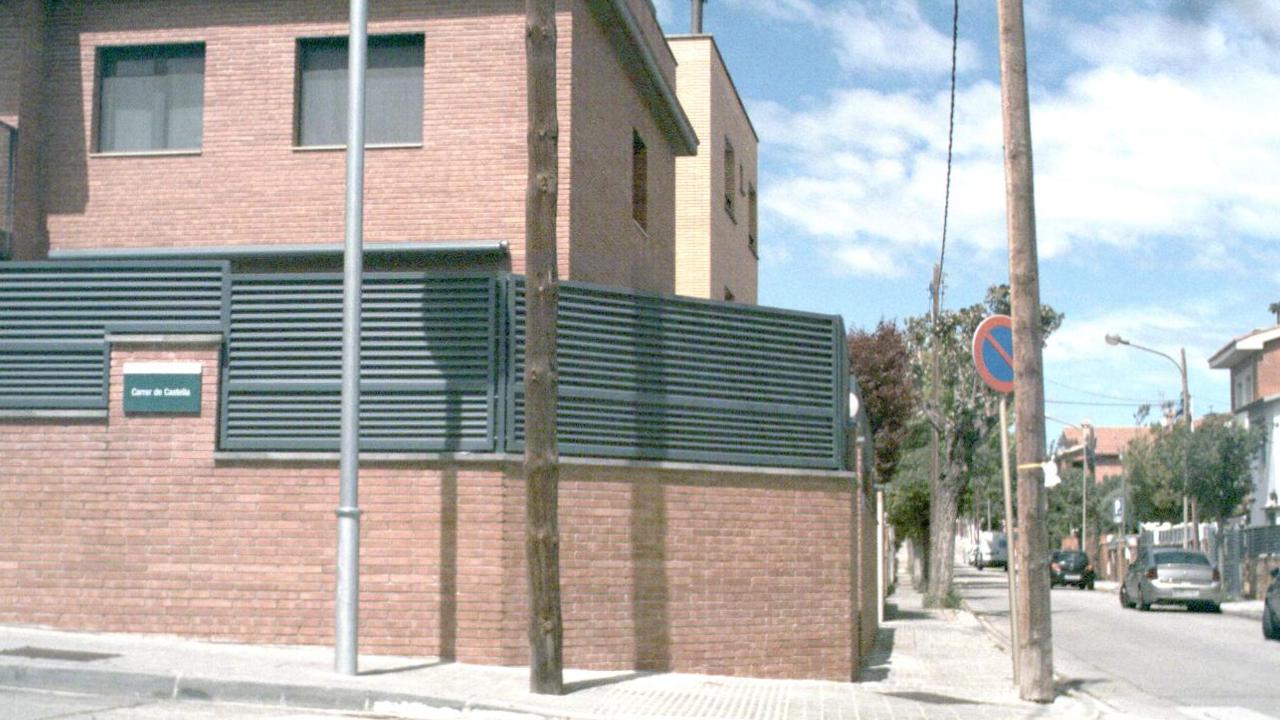} &
    \includegraphics[width=0.3\textwidth]{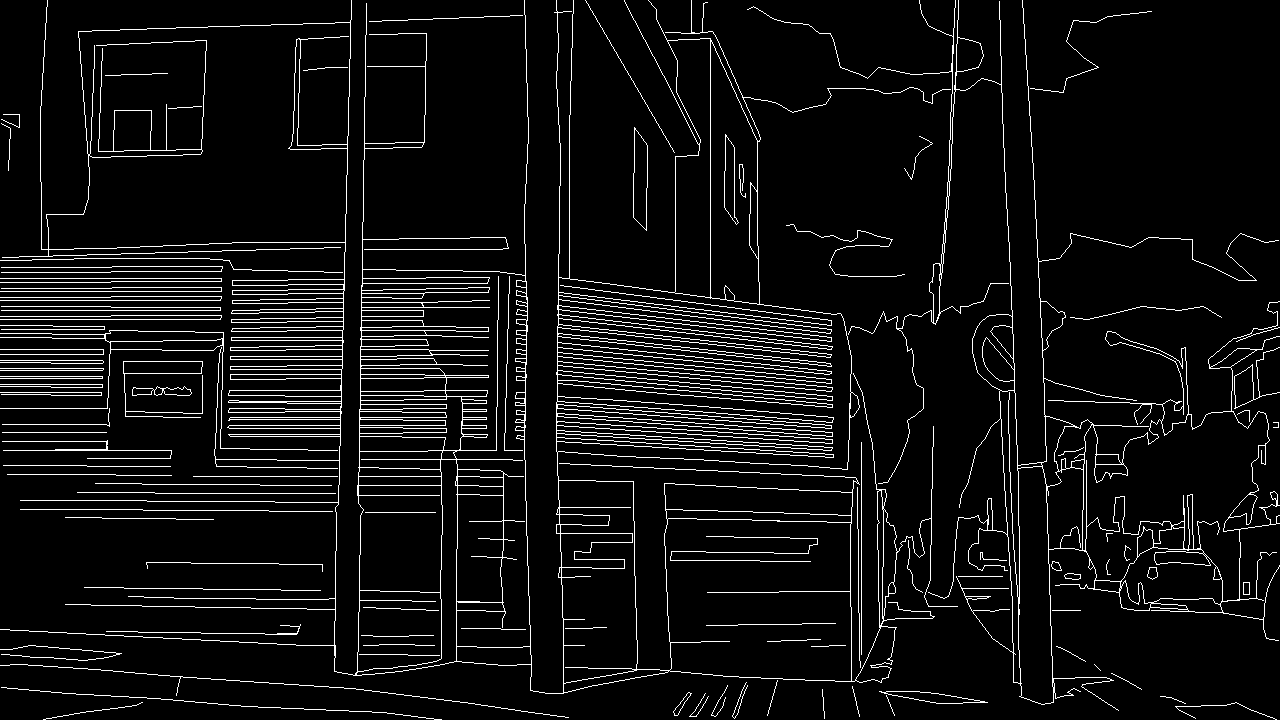}\\
    (a) image & (b) ground truth \\
    \includegraphics[width=0.3\textwidth]{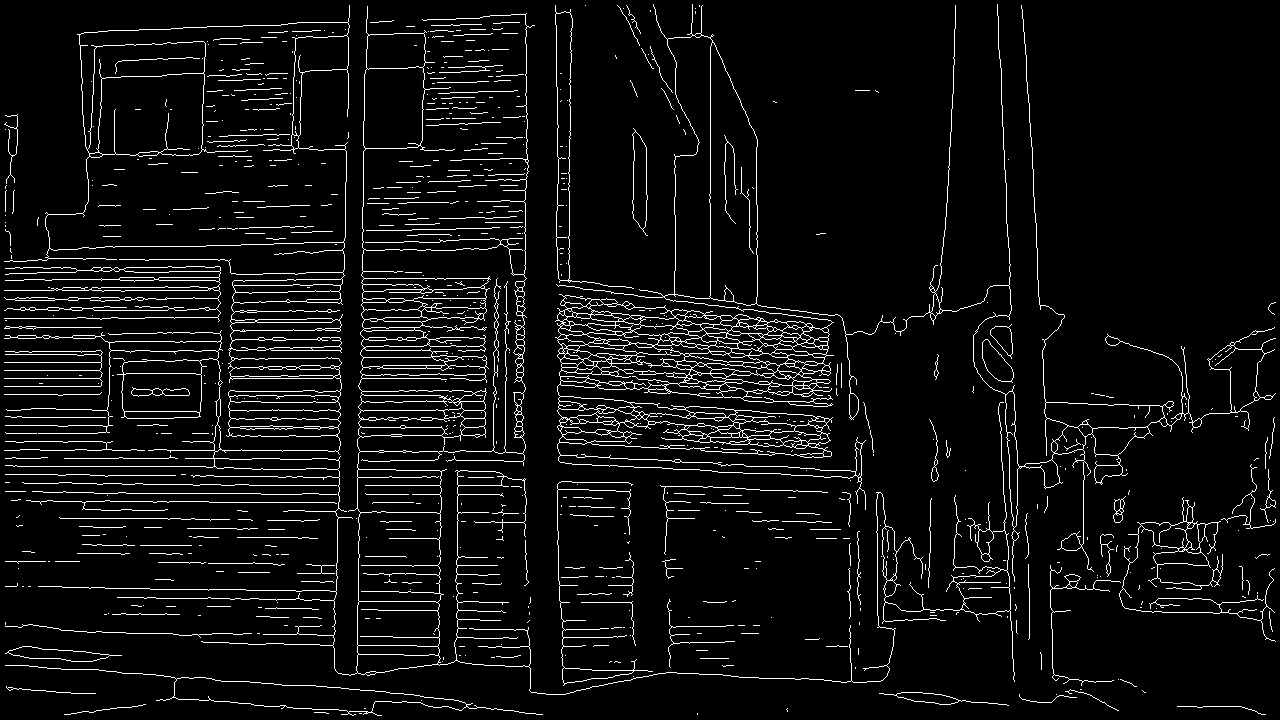} &
    \includegraphics[width=0.3\textwidth]{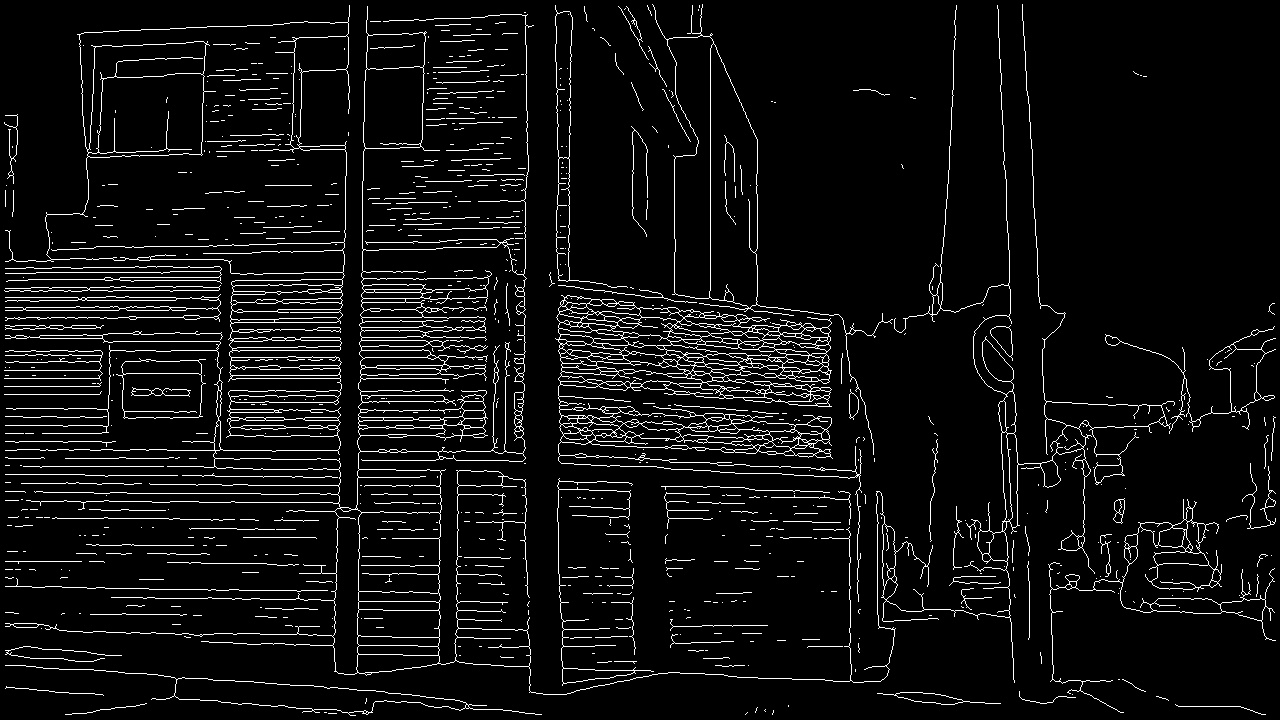} \\
    (c) HED & (d) RCF \\
    \includegraphics[width=0.3\textwidth]{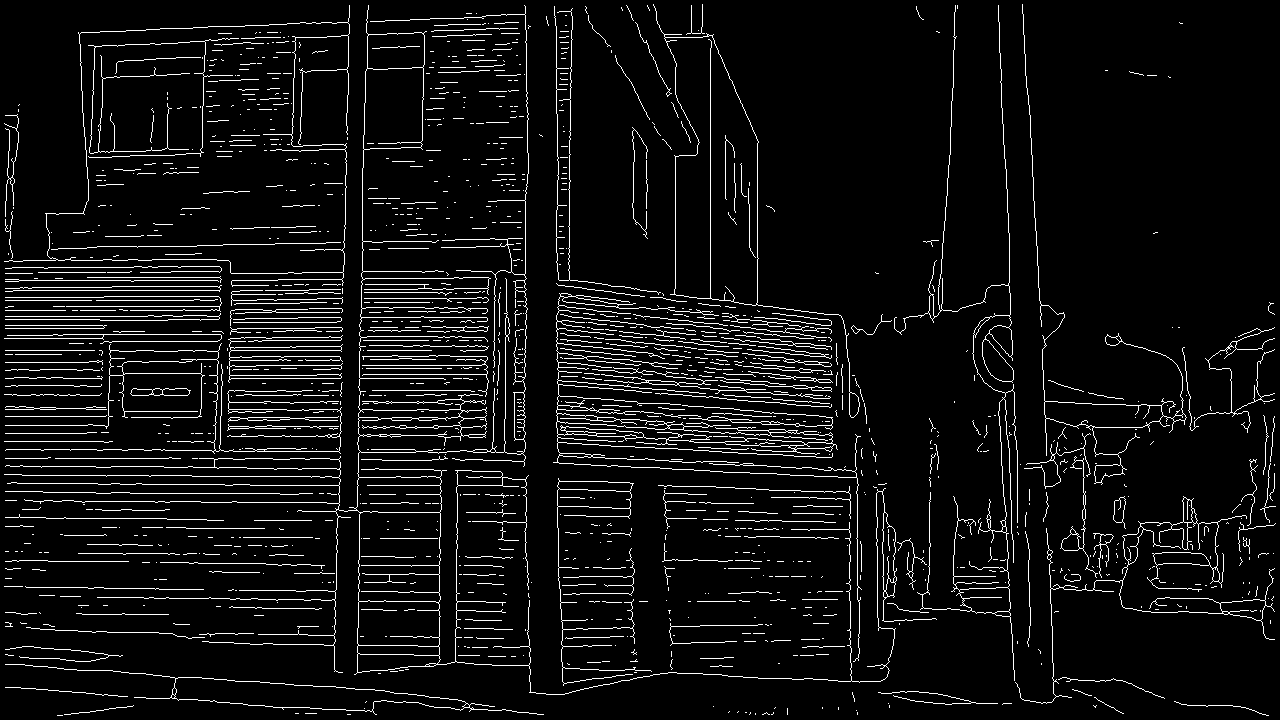} &
    \includegraphics[width=0.3\textwidth]{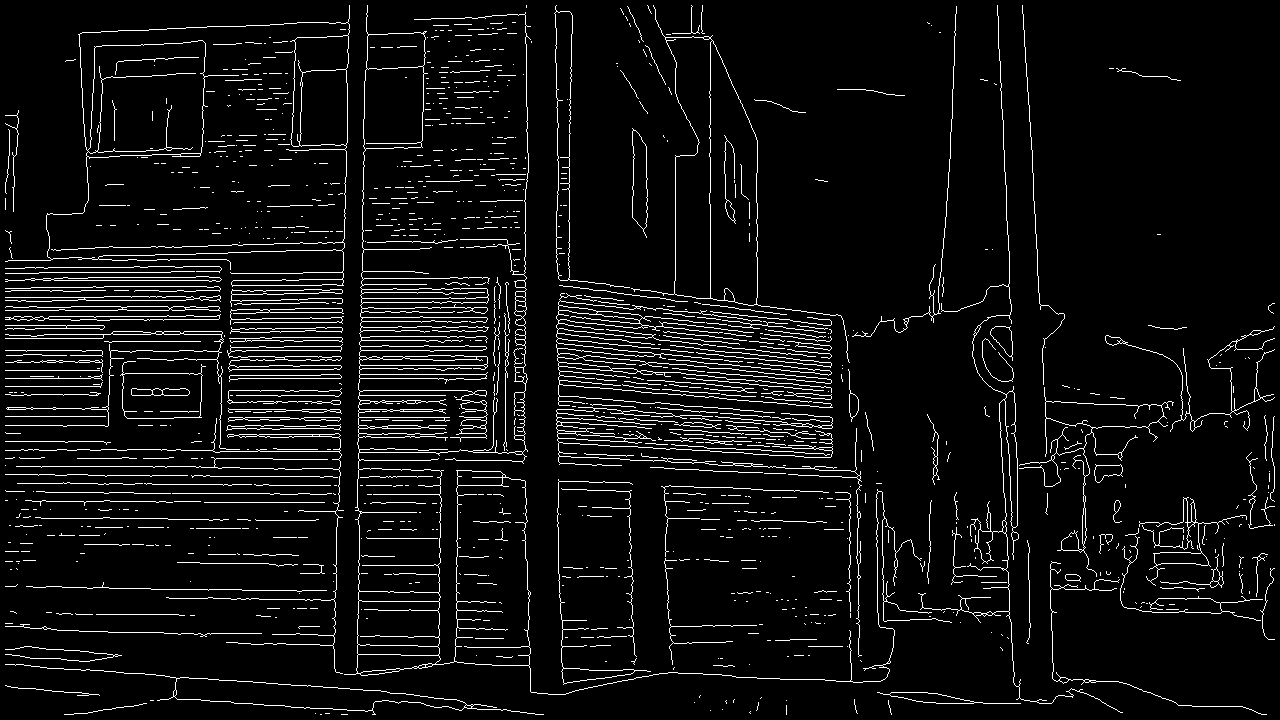} \\
    (e) BDCN & (f) CATS \\
    \includegraphics[width=0.3\textwidth]{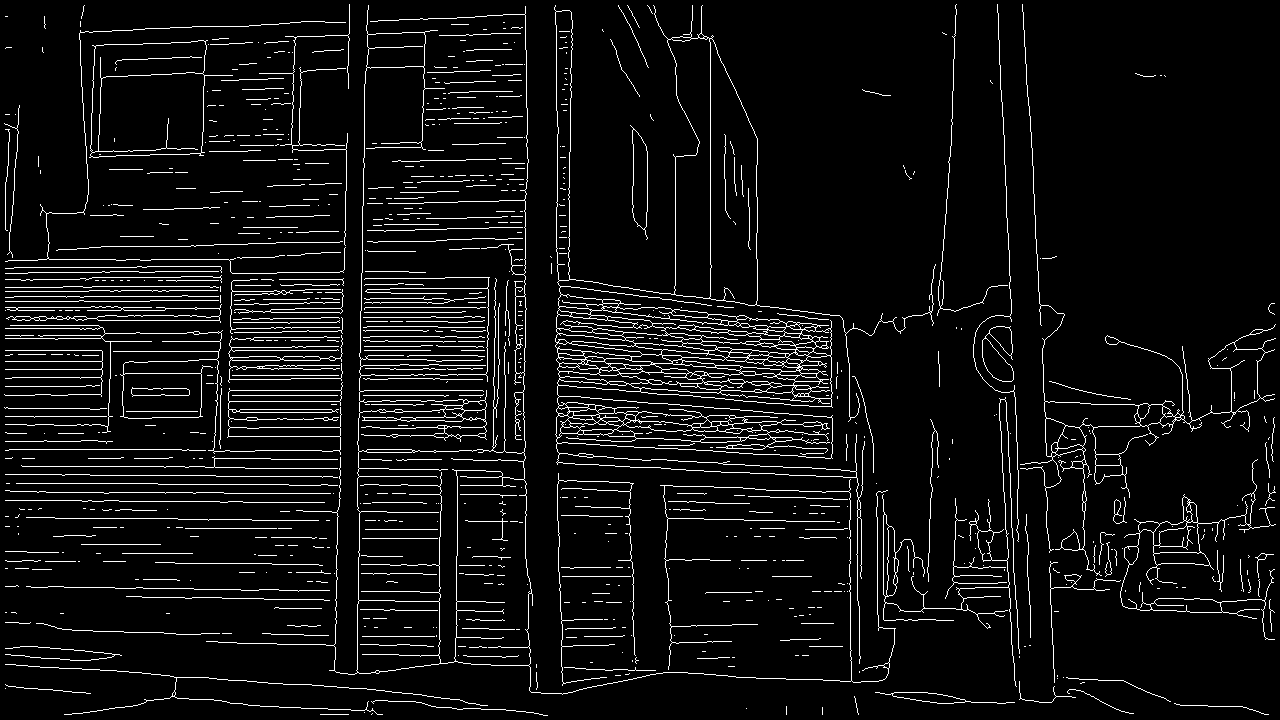} &
    \includegraphics[width=0.3\textwidth]{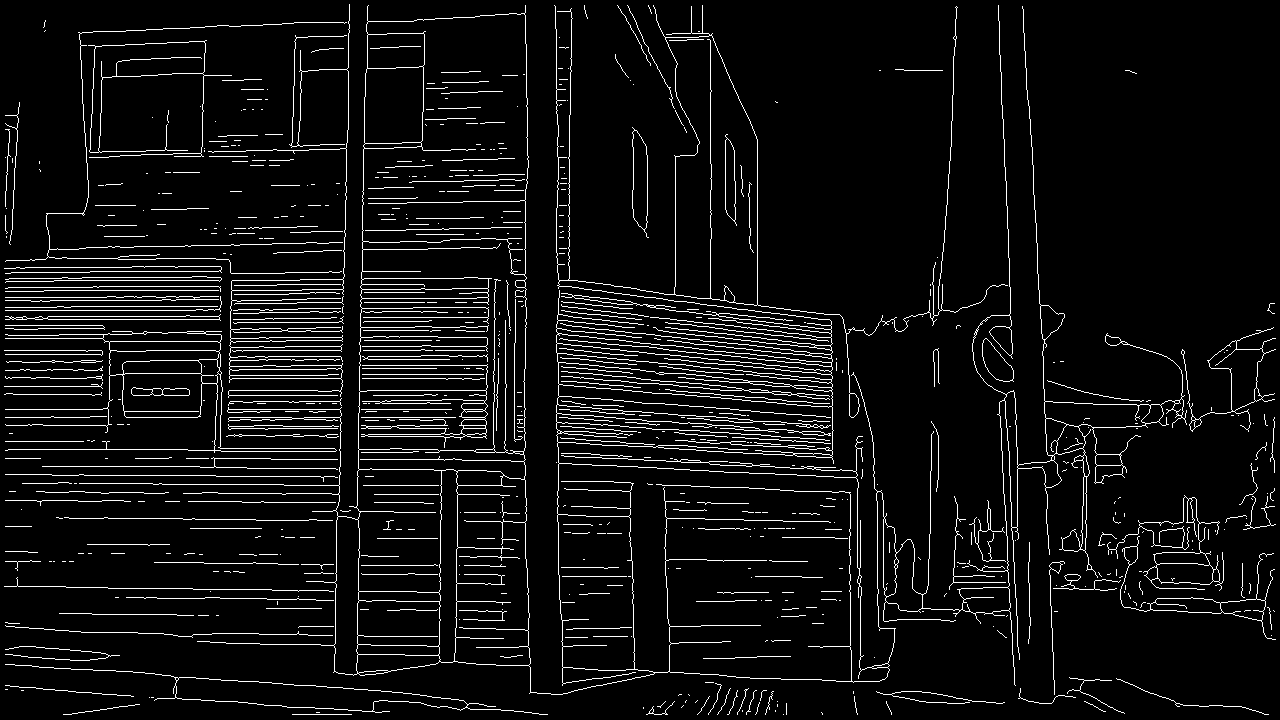} \\
   (g) DexiNed & (h) msmsfnet \\
  \end{tabular}
  \caption{Edge detection results computed by different edge detectors in a test image of BIPEDv2 dataset.}
  \label{biped_img}
  \end{figure}

The edge maps computed by different methods in an image from the test set of BIPEDv2 dataset are displayed in figure~\ref{biped_img}. 
From figure~\ref{biped_img} we can see that the edge maps computed by HED, RCF and DexiNed are noisy. 
Many false edges are detected by them. 
Compared to BDCN and CATS, fewer false edges are detected by our proposed method. 
Compared to the edge maps computed by other methods, the edge map computed by our method is cleaner and closer to the edge ground truth.

\subsection{BSDS500 dataset}
\label{sec_bsds}
The Berkeley Segmentation Dataset and Benchmark (BSDS500) dataset~\cite{bsds500} is one of the most commonly used dataset to evaluate the performance of edge detection algorithms. 
It contains 200 images for training, 100 images for validation and 200 images for testing. 
Every image in the dataset has multiple annotations (on average 5 annotations for each image) of the boundaries. 
Following previous work~\cite{rcf_pami, bdcn}, both training and validation images are augmented and mixed with the flipped VOC Context dataset~\cite{flippedvoc} to form the training dataset. 
As each image has been annotated by multiple subjects, special care should be taken to handle the disconsistency among them. 
Following previous work~\cite{rcf_pami, bdcn}, the edge ground truth for each training image is obtained as follows: 
a pixel is considered as a true edge pixel if at least three annotators label it as an edge pixel, 
and is considered as a non-edge pixel if none of the annotators has labeled it as an edge pixel; 
pixels that have been labeled as edges by fewer than three annotators are ignored during training. 
During testing, we adopt the multi-scale testing technique~\cite{rcf_pami, bdcn, cats} to improve the performance of each model. 
Specifically, each image is resized to three different scales (50$\%$, 100$\%$ and 150$\%$ of the original scale) and fed into the network separately. 
The edge predictions are then resized to the original scale of the input image and averaged to produce the final edge prediction.

\begin{figure}[h]
  \centering
  \begin{tabular}{c}
    \includegraphics[width=0.48\textwidth]{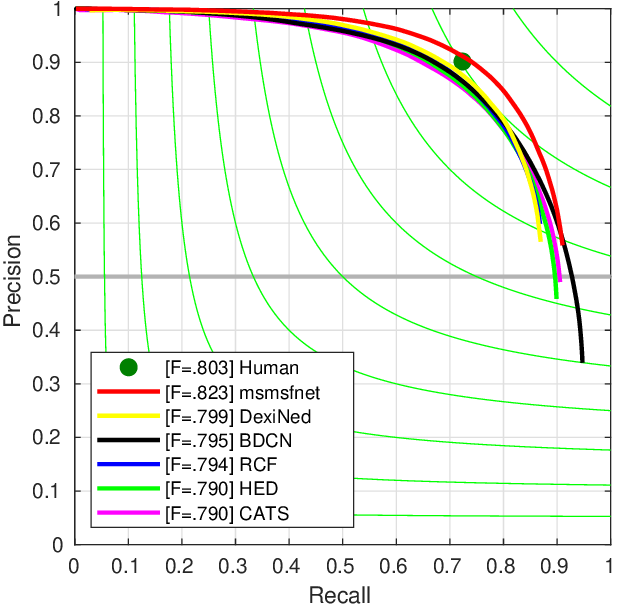}\\
  \end{tabular}
  \caption{The precision-recall curves computed by different methods in the test set of BSDS500 dataset.}
  \label{bsds_pr_curve}
  \end{figure}

\begin{figure}[h]
  \centering
  \begin{tabular}{cccc}
    \includegraphics[width=0.18\textwidth]{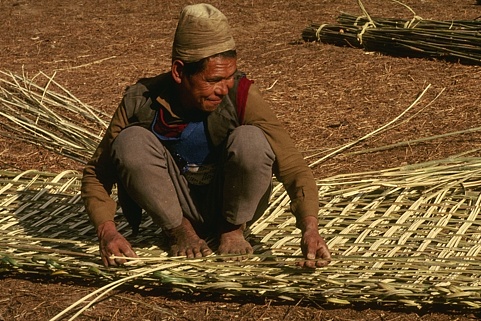} &
    \includegraphics[width=0.18\textwidth]{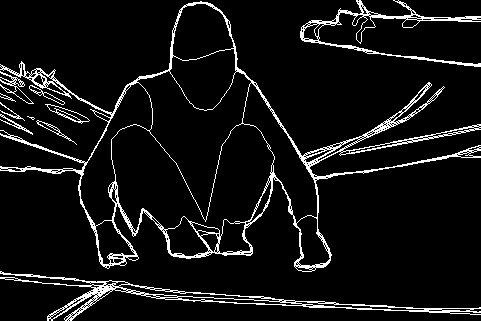} &
    \includegraphics[width=0.18\textwidth]{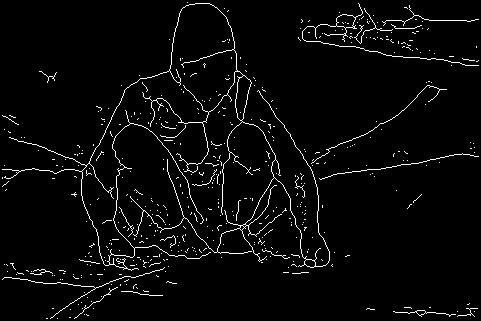} &
    \includegraphics[width=0.18\textwidth]{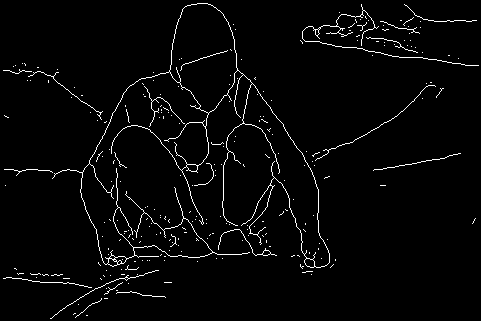} \\
    (a) image & (b) ground truth & (c) HED & (d) RCF \\
    \includegraphics[width=0.18\textwidth]{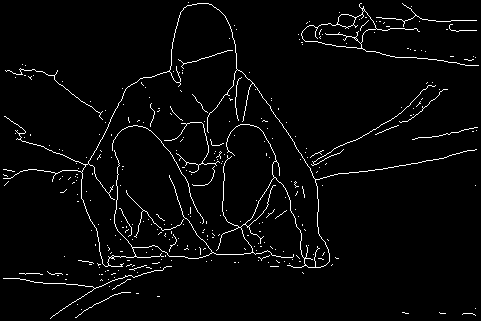} &
    \includegraphics[width=0.18\textwidth]{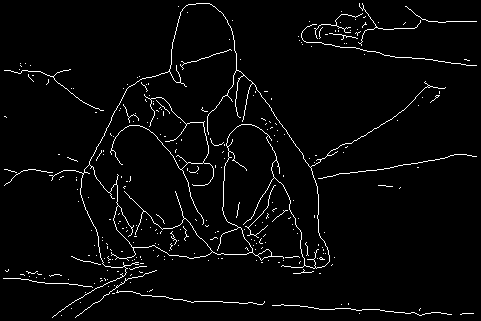} &
    \includegraphics[width=0.18\textwidth]{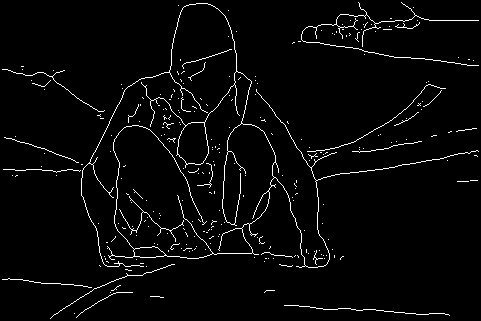} &
    \includegraphics[width=0.18\textwidth]{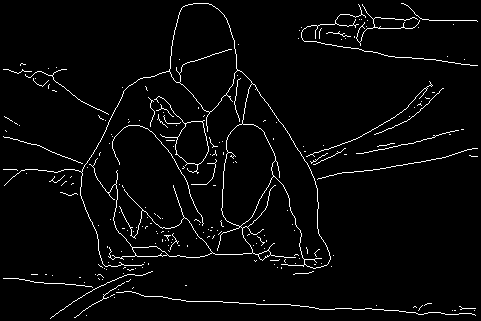} \\
    (e) BDCN & (f) CATS & (g) DexiNed & (h) msmsfnet \\
  \end{tabular}
  \caption{Edge detection results computed by different methods in a test image of BSDS500 dataset.}
  \label{bsds_img}
  \end{figure}

Table~\ref{f1_bsds} shows the quantitative comparisons of different methods in the test set of BSDS500 dataset. 
It is clearly shown that the proposed msmsfnet achieves much higher scores than state-of-the-art edge detectors under the same experimental settings. 
The ODS F1-score achieved by our method is at lease 2 percent higher than existing methods. 
In particular, the ODS F1-score computed by our method is 2 percent higher than humans, which shows that the perception ability of our model is stronger than humans even without the need to pre-train the model on the ImageNet dataset. 
The Average Precision computed by our method is at least 1.7 percent higher than existing methods. 
The performance gains obtained by our method in the BSDS500 dataset indicate the strong multi-scale representation capability of our model. 
Precision-recall curves computed by different methods are displayed in figure~\ref{bsds_pr_curve}. 
The edge maps computed by different methods in an image from the test set of BSDS500 dataset are shown in figure~\ref{bsds_img}. 
It is clearly shown that the edge map computed by our method is closer to the edge ground truth. 
More true edges are detected with fewer false detections. 
In contrast, the edges detected by BDCN, CATS and DexiNed tend to be fragmented, while many false edges are detected by HED.

\begin{table}
  \centering
  \caption{Quantitative comparisons of different algorithms on the test set of  BSDS500 dataset.}
  \label{f1_bsds}
  \setlength{\tabcolsep}{3pt}
  \begin{tabular}{c c c c}
    \toprule
    Methods & ODS F1-score & OIS F1-score & AP \\
    \midrule
    HED & 0.798 & 0.815 & 0.825 \\
    \midrule
    RCF & 0.798 & 0.815 & 0.815 \\
    \midrule
    BDCN & 0.803 & 0.821 & 0.837 \\
    \midrule
    CATS & 0.801 & 0.820 & 0.841 \\
    \midrule
    DexiNed & 0.799 & 0.818 & 0.810 \\
    \midrule
    msmsfnet & 0.823 & 0.840 & 0.858 \\
    \bottomrule
    \end{tabular}
\end{table}

\subsection{NYUDv2 dataset}
The NYU-Depth V2 (NYUDv2) dataset~\cite{nyudv2} contains 1449 pairs of RGB and depth images, which are well aligned, densely labeled and made into the same size. 
These images have been splitted into three parts: 381 images for training, 414 images for validation and 654 images for testing. 
The images in the dataset are video sequences of various indoor scenes and recorded by both RGB and Depth camera from Microsoft Kinect. 
Following previous work~\cite{rcf_pami, bdcn}, both training and validation images are augmented and used to train the models. 
Each image in the training and validation set is horizontally flipped and rotated to four different angles.
Each model has been trained twice: one trained using RGB images and one trained using HHA features. 
We compare the performance of different methods using three types of edge predictions produced by each model: 
one produced by the model trained on RGB images, 
one produced by the model trained on HHA features, 
and one produced by averaging the two previous predictions. 
During evaluation, the maximum tolerance allowed to match the edge predictions with the edge ground truth correctly is increased from 0.0075 to 0.011 as done in previous work~\cite{hed_ijcv, rcf_pami, bdcn}. 
Quantitative comparisons of different methods in the test set of NYUDv2 dataset are displayed in table~\ref{f1_nyud}.
\begin{table}
  \centering
  \caption{Quantitative comparisons of different algorithms on the test set of  NYUDv2 dataset.}
  \label{f1_nyud}
  \setlength{\tabcolsep}{3pt}
  \begin{tabular}{c c c c}
    \toprule
    Methods & ODS F1-score & OIS F1-score & AP \\
    \midrule
    $\text{HED}_{\text{RGB}}$ & 0.704  & 0.720 & 0.714 \\
    \midrule
    $\text{RCF}_{\text{RGB}}$ & 0.715  & 0.731 & 0.717 \\
    \midrule
    $\text{BDCN}_{\text{RGB}}$ & 0.709 & 0.728 &  0.692 \\
    \midrule
    $\text{CATS}_{\text{RGB}}$ & 0.713 & 0.730 & 0.721 \\
    \midrule
    $\text{DexiNed}_{\text{RGB}}$ & 0.721 & 0.734 & 0.650 \\
    \midrule
    $\text{msmsfnet}_{\text{RGB}}$ & 0.732 & 0.747 & 0.744\\
    \midrule
    $\text{msmsfnet}_{\text{RGB}}^{aug+multi}$ & 0.747 & 0.764 & 0.767\\
    \midrule
    \midrule
    $\text{HED}_{\text{HHA}}$ & 0.674 & 0.688 & 0.680\\
    \midrule
    $\text{RCF}_{\text{HHA}}$ & 0.685 & 0.701 & 0.680 \\
    \midrule
    $\text{BDCN}_{\text{HHA}}$ & 0.677 & 0.694 & 0.692\\
    \midrule
    $\text{CATS}_{\text{HHA}}$ & 0.678 & 0.695 & 0.679\\
    \midrule
    $\text{DexiNed}_{\text{HHA}}$ & 0.669 & 0.686 & 0.612 \\
    \midrule
    $\text{msmsfnet}_{\text{HHA}}$ & 0.704 & 0.721 & 0.727 \\
    \midrule
    $\text{msmsfnet}_{\text{HHA}}^{aug+multi}$ & 0.720 & 0.736 & 0.744 \\
    \midrule
    \midrule
    $\text{HED}_{\text{RGB+HHA}}$ & 0.716 & 0.739 & 0.758 \\
    \midrule
    $\text{RCF}_{\text{RGB+HHA}}$ & 0.730  & 0.753 & 0.764 \\
    \midrule
    $\text{BDCN}_{\text{RGB+HHA}}$ & 0.728 & 0.751 & 0.774\\
    \midrule
    $\text{CATS}_{\text{RGB+HHA}}$ & 0.726 & 0.748 & 0.773 \\
    \midrule
    $\text{DexiNed}_{\text{RGB+HHA}}$ & 0.743 & 0.757 & 0.735\\
    \midrule
    $\text{msmsfnet}_{\text{RGB+HHA}}$ & 0.746 & 0.767 & 0.789 \\
    \midrule
    $\text{msmsfnet}_{\text{RGB+HHA}}^{aug+multi}$ & 0.763 & 0.783 & 0.803 \\
    \bottomrule
    \end{tabular}
  \end{table}

As shown in table~\ref{f1_nyud}, our model achieves the highest scores in all experimental settings. 
When all models are trained using only RGB images, the ODS F1-score computed by our method is at least 1.1 percent higher than those computed by existing methods. 
The Average Precision computed by our method is at least 2.3 percent higher than those computed by existing methods. 
When only HHA features are used to train the model, the ODS F1-score computed by our method is at least 1.9 percent higher. 
The improvements on the Average Precision is much larger, at least 3.5 percent higher than state-of-the-art methods. 
For the edge predictions computed by combining the information from both RGB images and HHA features, the improvements on the ODS F1-score is relatively minor, only 0.3 percent higher than the best result (computed by DexiNed). 
However, the Average Precision computed by our method is 5.4 percent higher than that of DexiNed. 
The ODS F1-score computed by our method is at least 1.6 percent higher than the rest methods. 
Similar improvements have been done on the Average Precision, at least 1.5 percent higher than the rest methods. 

We also show that by using more complicated data augmentation technique and multi-scale testing, the performance of the proposed method (denoted as $\text{msmsfnet}_{*}^{aug+multi}$) can be improved further.
Specifically, each image in the training and validation set is horizontally flipped, vertically flipped and flipped in both directions.
Then, all images are rescaled to five different scales: $50\%$, $80\%$, $100\%$, $120\%$, and $150\%$.
The multi-scale testing technique is the same as the one used in Subsection ~\ref{sec_bsds}.
As shown in table ~\ref{f1_nyud}, significant improvements have been done by the proposed msmsfnet when using more complicated data augmentation technique and multi-scale testing.

\begin{figure}[h]
  \centering
  \begin{tabular}{ccc}
    \includegraphics[width=0.28\textwidth]{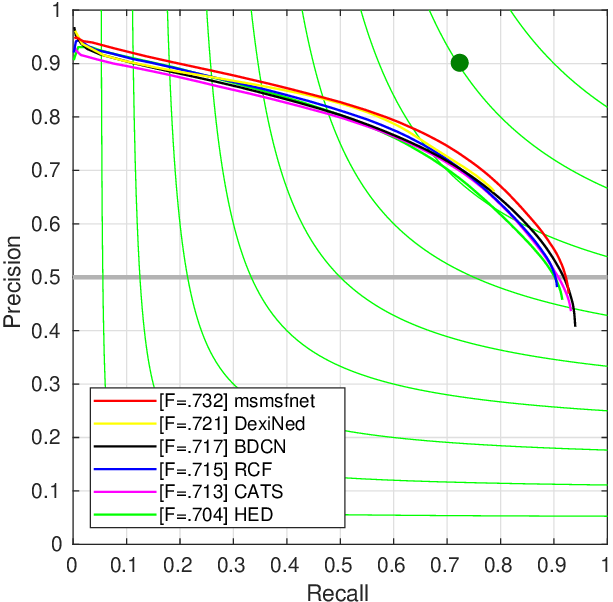} &
    \includegraphics[width=0.28\textwidth]{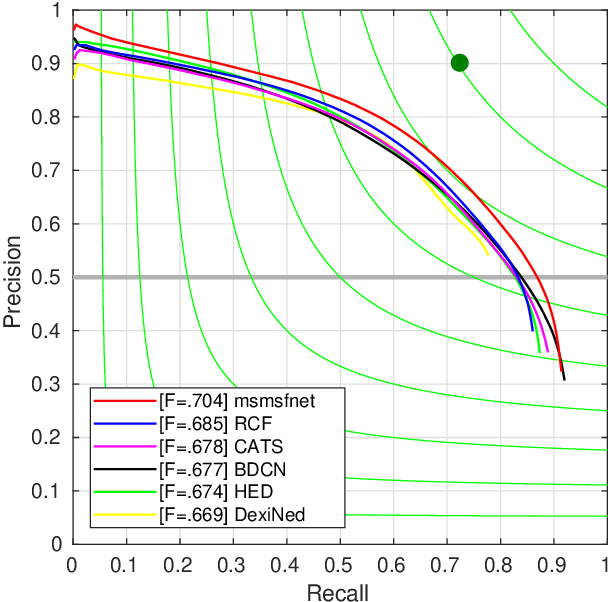} &
    \includegraphics[width=0.28\textwidth]{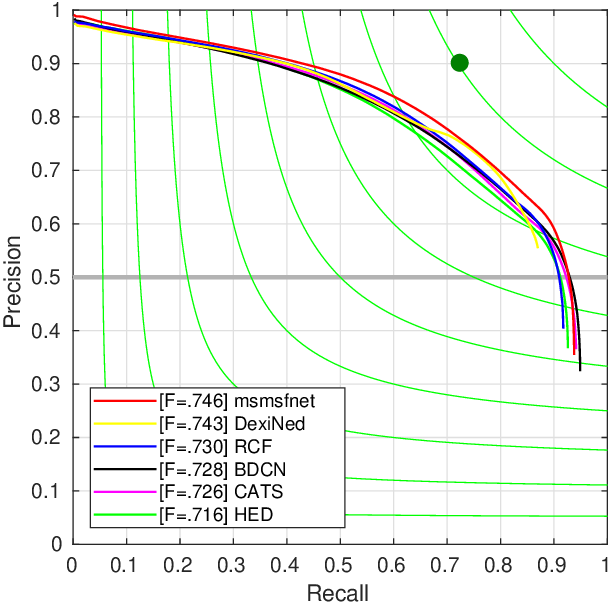} \\
    (a) RGB & (b) HHA & (c) RGB$+$HHA \\
  \end{tabular}
  \caption{The precision-recall curves computed by different methods in the NYUDv2 dataset.}
  \label{nyud_pr_curve}
  \end{figure}

\begin{figure}[h]
  \centering
  \begin{tabular}{cccc}
    \includegraphics[width=0.18\textwidth]{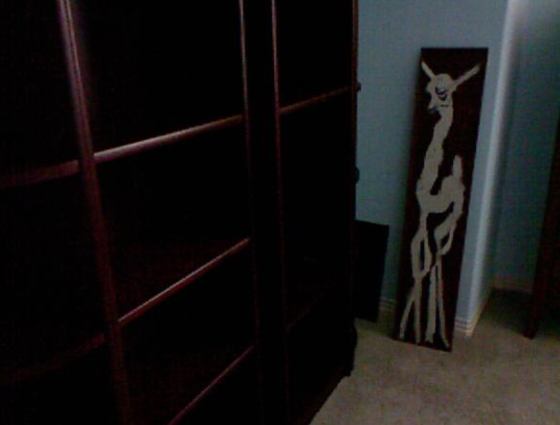} &
    \includegraphics[width=0.18\textwidth]{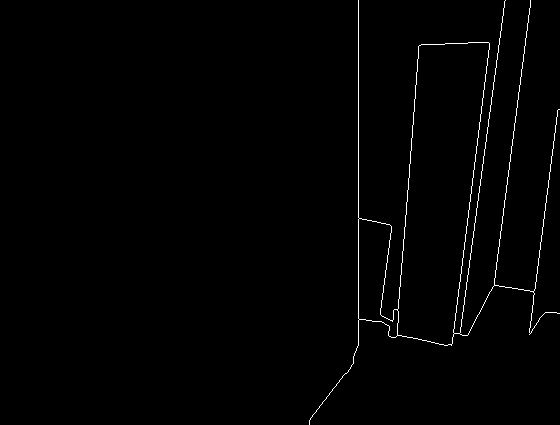} &
    \includegraphics[width=0.18\textwidth]{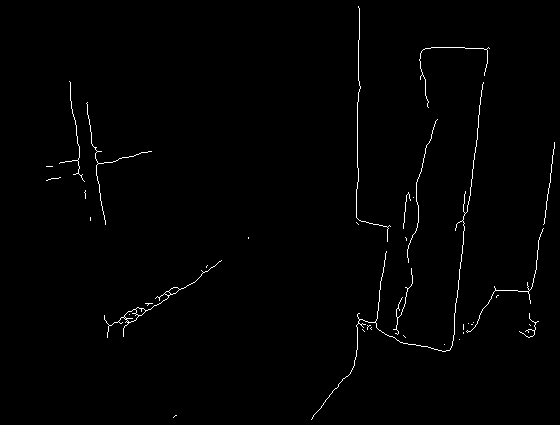} &
    \includegraphics[width=0.18\textwidth]{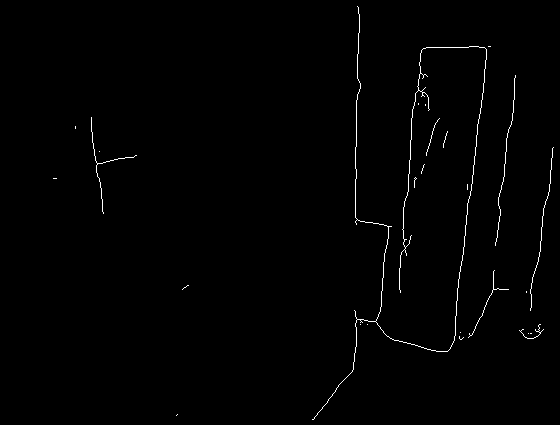} \\
    (a) RGB image & (b) ground truth & (c) HED & (d) RCF \\
    \includegraphics[width=0.18\textwidth]{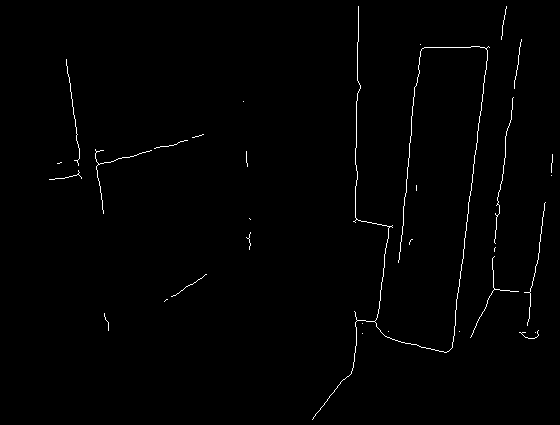} &
    \includegraphics[width=0.18\textwidth]{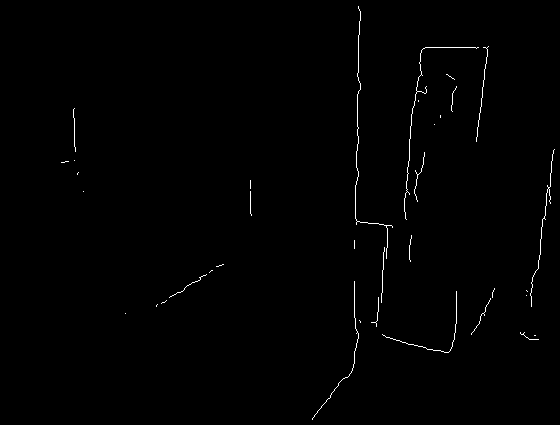} &
    \includegraphics[width=0.18\textwidth]{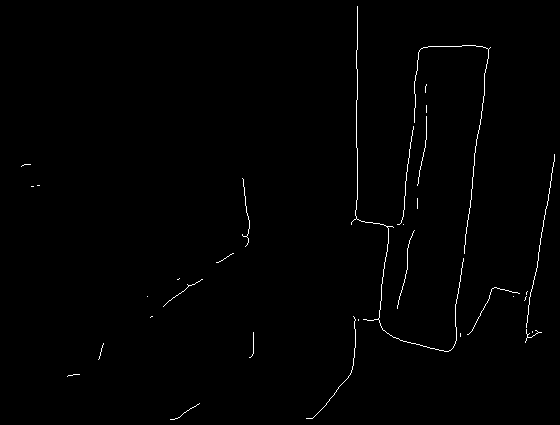} &
    \includegraphics[width=0.18\textwidth]{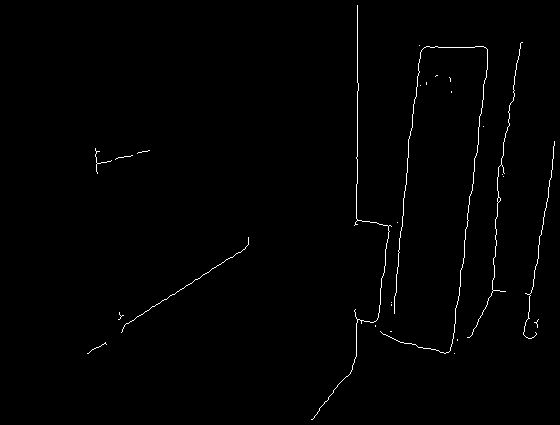} \\
    (e) BDCN & (f) CATS &    (g) DexiNed & (h) msmsfnet \\
  \end{tabular}
  \caption{Edge detection results computed by different methods in a test image (RGB) of the NYUDv2 dataset.}
  \label{nyud_rgb_img}
  \end{figure}

\begin{figure}[h]
  \centering
  \begin{tabular}{cccc}
    \includegraphics[width=0.18\textwidth]{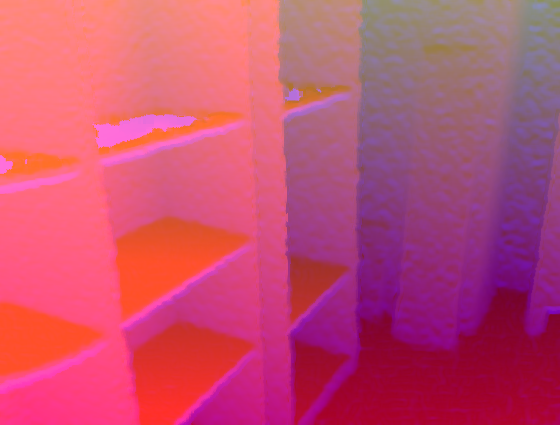} &
    \includegraphics[width=0.18\textwidth]{images/nyud_png/gt/img_6149.png} &
    \includegraphics[width=0.18\textwidth]{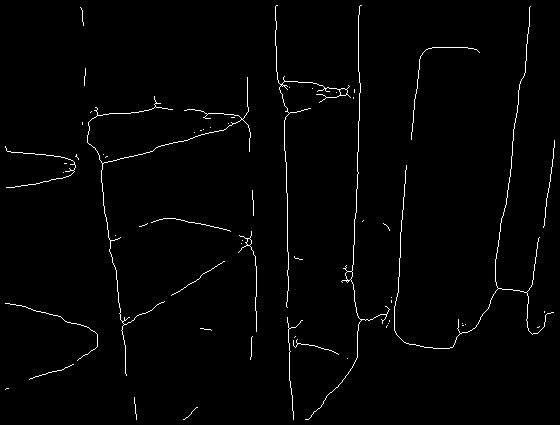} &
    \includegraphics[width=0.18\textwidth]{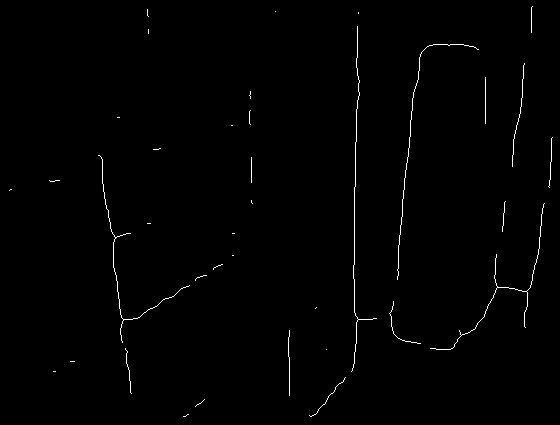} \\
    (a) HHA feature & (b) ground truth &
    (c) HED & (d) RCF\\
    \includegraphics[width=0.18\textwidth]{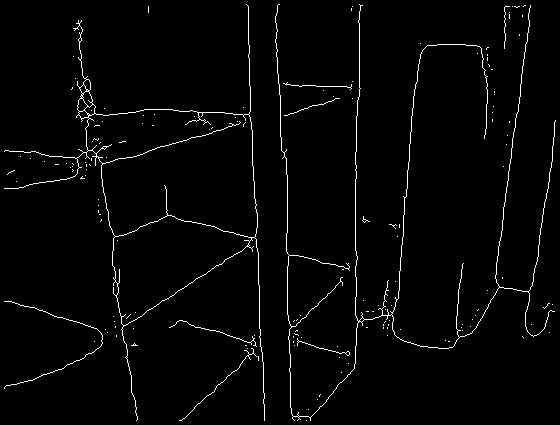} &
    \includegraphics[width=0.18\textwidth]{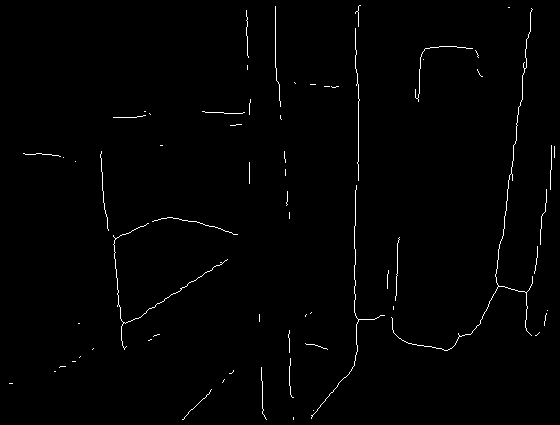} &
    \includegraphics[width=0.18\textwidth]{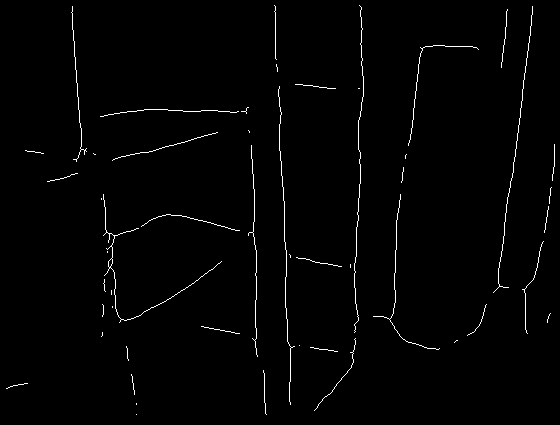} &
    \includegraphics[width=0.18\textwidth]{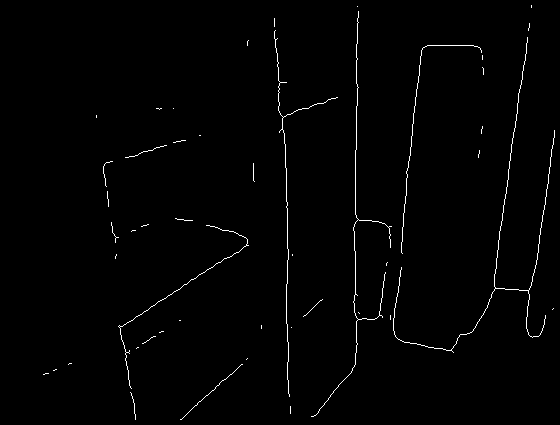} \\
    (e) BDCN & (f) CATS &
    (g) DexiNed & (h) msmsfnet \\
  \end{tabular}
  \caption{Edge detection results computed by different methods in a test image (HHA) of the NYUDv2 dataset.}
  \label{nyud_hha_img}
  \end{figure}

The precision-recall curves computed by different methods in all experimental settings are displayed in figure~\ref{nyud_pr_curve}. 
The edge maps computed by different methods in an image from the test set of the NYUDv2 dataset can be found in figure~\ref{nyud_rgb_img} and figure~\ref{nyud_hha_img}. 
Figure~\ref{nyud_hha_img} displays the results computed by different methods using only HHA features. 
It can be seen from these figures that the edges detected by our method are more complete and fewer false edges are detected.

\subsection{Comparison in SAR images}
As described in Section~\ref{introduction}, the statistics of SAR images differs greatly from that of natural images. 
Therefore, fine-tuning the model from the pre-trained weights of the backbone network on the ImageNet dataset is unlikely to improve the edge detection accuracy in real SAR images even in the presence of a real SAR dataset.
What's worse, there is still no real SAR dataset to train CNN models for edge detection.
GRHED~\cite{grhed} proposed that we could train CNN models on simulated datasets for edge detection in real SAR images in the absence of a real SAR dataset. 
It proposed to simulate SAR images by multiplying natural images with 1-look speckle noise. 
However, strong differences still exist between simulated SAR and real SAR images, making models trained directly using simulated SAR images inefficient for real SAR images~\cite{grhed}. 
To tackle this issue, GRHED proposed to combine a ratio operator with fully convolutional layers to handle the differences between simulated SAR and real SAR images. 
This special design, which enables models trained using simulated SAR images to perform efficient edge detection in real SAR images, makes the pre-trained weights of the backbone network on the ImageNet dataset useless because of the ratio operator injected before convolutional layers.

In the following, we demonstrate the efficiency of our proposed msmsfnet by comparing it with a traditional ratio operator GR~\cite{SARSIFT15} and state-of-the-art CNN-based edge detectors in both simulated SAR dataset and a 1-look real SAR image.  
Following the paradigm of GRHED~\cite{grhed}, we build GR-HED, GR-RCF, GR-BDCN, GR-CATS, GR-DexiNed, and GR-msmsfnet by combining a ratio operator GR~\cite{SARSIFT15} with HED~\cite{hed_ijcv}, RCF~\cite{rcf_pami}, BDCN~\cite{bdcn}, CATS~\cite{cats}, DexiNed~\cite{dexined} as well as our proposed msmsfnet. 
We simulate a 1-look SAR dataset with the augmented BSDS500 dataset~\cite{bsds500} and the flipped VOC Context dataset~\cite{flippedvoc} as described in ~\cite{rcf_pami} by multiplying the images with 1-look speckle noise~\cite{grhed}. 
We compare the performance of different methods in the two hundred 1-look  images which are simulated from the test set of BSDS500 and a 1-look real SAR image  (SLC data of Sentinel-1, ESA, IW mode, date: May 12, 2021, area: Paris, France). 

\begin{table}[h]
  \centering
  \caption{Comparison of different methods in the 200 1-look speckled optical images simulated using the test set of BSDS500.}
  \label{F1-score}
  \setlength{\tabcolsep}{6pt}
  \begin{tabular}{c c c c}
    \toprule
    Method & ODS & OIS & AP\\
    \midrule
    GR & 0.589 & 0.615 & 0.479 \\
    \midrule
    GR-HED & 0.695 & 0.711 & 0.716 \\
    \midrule
    GR-RCF & 0.693 & 0.708 & 0.708 \\
    \midrule
    GR-BDCN & 0.694 & 0.710 & 0.715 \\
    \midrule
    GR-CATS & 0.695 & 0.711 & 0.722 \\
    \midrule
    GR-DexiNed & 0.697 & 0.714 & 0.709 \\
    \midrule
    GR-msmsfnet & 0.700 & 0.717 & 0.724 \\
    \bottomrule
    \end{tabular}
\end{table}

Quantitative comparisons of different methods in the simulated SAR dataset are shown in Table~\ref{F1-score}. It is clearly shown in Table~\ref{F1-score} that our proposed msmsfnet achieves higher scores in all criteria, which demonstrates the efficiency of our proposed msmsfnet. Besides, the scores computed by the deep learning based edge detectors are significantly higher (around 10 percent higher in terms of ODS F1-score and OIS F1-score, and more than 20 percent higher in terms of AP) than those of GR~\cite{SARSIFT15}, which demonstrates the benefits of exploiting convolutional neural networks to improve the performance of edge detectors in images corrupted by strong noise. 

\begin{figure}[h]
  \centering
  \begin{tabular}{cccc}
    \includegraphics[width=0.18\textwidth]{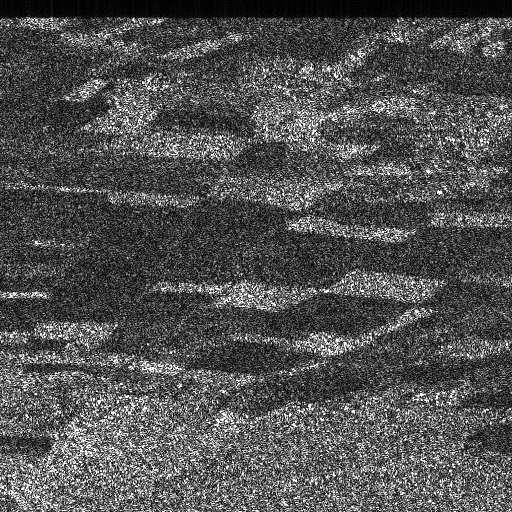} &
     \includegraphics[width=0.18\textwidth]{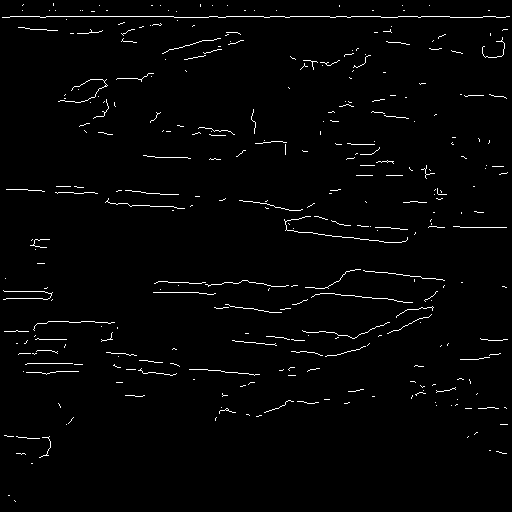} &
     \includegraphics[width=0.18\textwidth]{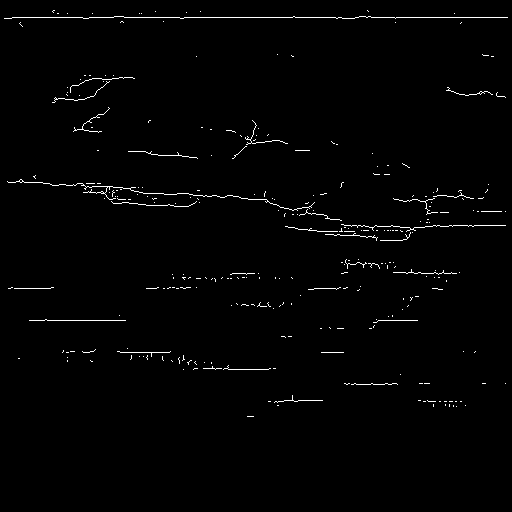} &
     \includegraphics[width=0.18\textwidth]{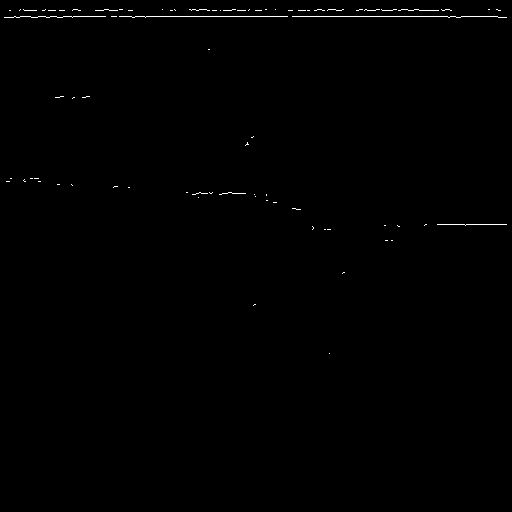} \\
     (a) Sentinel-1 image & (b) GR & (c) GR-HED & (d) GR-RCC \\
      \includegraphics[width=0.18\textwidth]{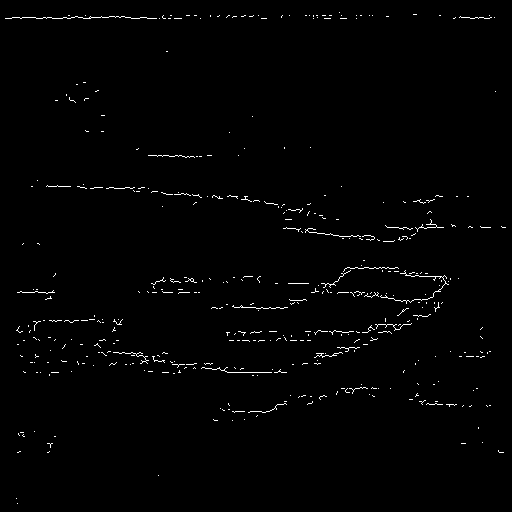} &
     \includegraphics[width=0.18\textwidth]{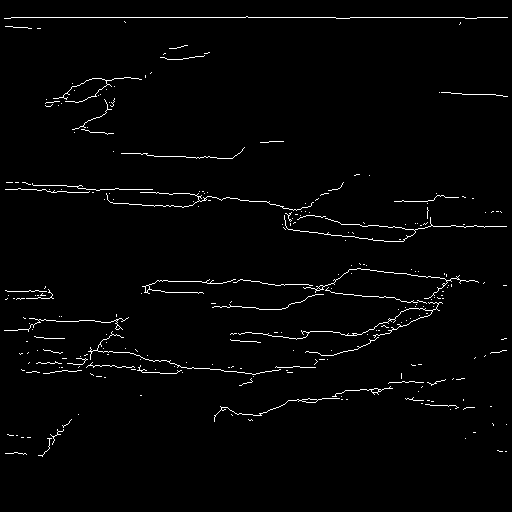} &
     \includegraphics[width=0.18\textwidth]{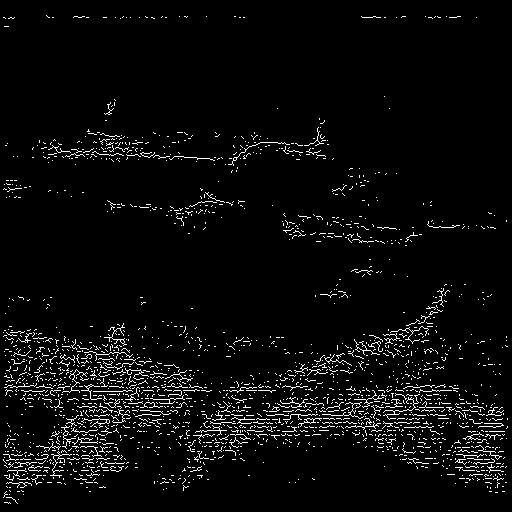} &
     \includegraphics[width=0.18\textwidth]{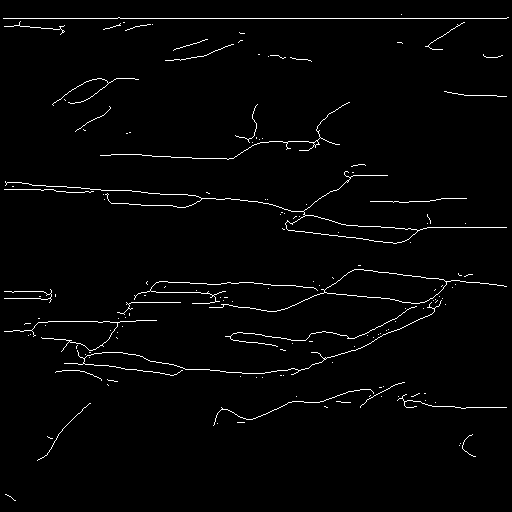} \\
     (e) GR-BDCN & (f) GR-CATS & (g) GR-DexiNed & (h) GR-msmsfnet \\
     \end{tabular}
     \caption{Edge detection results computed by different edge detection algorithms in a 1-look real SAR image (Sentinel-1) of size $512\times 512$ pixels.}
     \label{real_sar}
  \end{figure}

We further compare the performance of different algorithms in a 1-look real SAR image of size $512\times 512$ pixels. 
The edge maps computed by different algorithms are displayed in Figure~\ref{real_sar}. 
It is clearly shown in Figure~\ref{real_sar} that our proposed method GR-msmsfnet achieves the best edge detection results with more true edges detected and much fewer false detections. 
As shown in Figure~\ref{real_sar}-(h), the edges detected by our proposed GR-msmsfnet are more complete. 
Compared to the traditional ratio operator GR, GR-CATS also achieves better edge detection result with more complete edges and fewer false detections. 
However, the edge detection results computed by GR-HED, GR-RCF, GR-BDCN and GR-DexiNed in the real SAR image are very poor and worse than that computed by GR, although the scores computed by them in the simulated SAR dataset are significantly higher than those of GR. The efficiency of GR-msmsfnet and GR-CATS in the real SAR image shows that the strategy of combining a traditional ratio operator with fully convolutional neural networks successfully enables the models trained using simulated SAR images to work in real SAR images. In contrast, the poor edge detection results computed by GR-HED, GR-RCF, GR-BDCN and GR-DexiNed in the real SAR image poses an important issue of CNNs: its generalization capability. This is indeed worthy of study in future work.

\subsection{Pre-training and fine-tuning}
\label{pre_fine}
In order to demonstrate the efficiency of the proposed method, we show that our method is able to achieve competitive performance in the BSDS500 dataset when the pre-trained weights from the ImageNet dataset are used to initialize model's parameters.
To obtain the pre-trained weights on the ImageNet dataset, we modify out network architecture for image classification.
We add a max-pooling layer, a global average pooling layer and a 1000-way fully connected layer after the last msmsfblock so that it can perform the task of image classification on the ImageNet dataset.
All the side layers are removed.
The modified model is trained using the training dataset of the ImageNet-1k dataset, which contains around 1.28 million images.
All training Images (RGB) are randomly resized and then cropped into size of $224\times 224$ pixels to be used as the input of the network.
Randomly horizontal flipping is used for data augmentation. 
We train our model using the SGD optimizer and use the cross-entropy loss as the loss function.
During training, the learning rate is initially set to 0.01, and divided by 10 every 60 epochs. 
The training batch size is set to 256, momentum is set to 0.9, and the weight decay is set to 0.0001.
Our model reaches its best performance on the validation set of the ImageNet dataset after training for 145 epochs.
It achieves $70.91\%$ top-1 accuracy, which is slightly lower (less than one percent) than the VGG net and around five percent lower than the ResNet-50 net.
It should be noted that our aim here is just to obtain the pre-trained weights from the ImageNet dataset which will be used to fine-tune our model for edge detection.
The performance of our model may be improved if a better training strategy is used.


After obtaining the pre-trained weights, our model is fine-tuned for edge detection on the the BSDS500 dataset.
During fine-tuning, we find it beneficial to adopt the fusion strategy of BDCN. 
Thus, we modify our network architecture following the fusion strategy of BDCN to obtain better performance.
During fine-tuning, our experimental settings are the same as ~\cite{rcf_pami, bdcn}.
The training dataset and the loss function is the same as those used in Subsection~\ref{sec_bsds}.
Following previous work ~\cite{hed_ijcv, rcf_pami, bdcn}, we use SGD optimizer to fine-tune our model for edge detection.
The learning rate is initially set to $10^{-6}$, and divided by 10 after 12 epochs. 
Momentum is set to 0.9 and weight decay is set to 0.0002.
We train our model for 20 epochs and the best performance (msmsfnet$^{pre-training}$) is reported in table ~\ref{bsds_pre}.
The performance obtained by msmsfnet with random initialization (msmsfnet$^{random}$) is also reported.

We compare the performance of our model with BDCN, which reported the best performance  on the BSDS500 dataset among state-of-the-art CNN-based edge detectors, 
and EDTER, a transformer-based edge detector which reports the best performance for edge detection on the BSDS500 dataset.

For BDCN, four results are reported. 
Two of them are copied from the paper: BDCN$^{paper}_{VGG}$ and BDCN$^{paper}_{ResNet}$, which use the VGG net  and ResNet net as the backbone network, respectively.
The third one is obtained through our evaluation of BDCN$_{VGG}$  using the pre-trained weights provided by the author, namely BDCN$^{testing}_{VGG}$.
The fourth one, BDCN$^{random}_{VGG}$, is obtained by training BDCN$_{VGG}$ with random initialization.

For EDTER, we report three results.
One is copied from the paper (EDTER$^{paper}$), one is obtained through our re-training of EDTER following the training settings in the paper (EDTER$^{re-training}$), 
and the third one is obtained by training EDTER from scratch (EDTER$^{random}$).
In order to get the best performance achievable by EDTER with random initialization, we train EDTER using both Adam optimizer and SGD optimizer.
In both training settings, the model is trained for 30 epochs, and the learning rate is divided by 10 after 20 epochs.
the weight decay is set to $10^{-12}$.
The mini-batch size for training is 6.
Three initial learning rates are used for the Adam optimizer, namely $10^{-3}$, $10^{-4}$, and $10^{-5}$.
Five initial learning rates are used for the SGD optimizer: $10^{-3}$, $10^{-4}$, $10^{-5}$, $10^{-6}$, and $10^{-7}$.
The best performance obtained by EDTER are reported (see table~\ref{bsds_pre}).

\begin{table}
  \centering
  \caption{Quantitative comparison between BDCN, EDTER and the proposed msmsfnet on the test set of BSDS500 dataset. The pre-trained weights for BDCN and msmsfnet are obtained by pre-training the backbone network on the ImageNet-1k dataset, while the pre-trained weights for EDTER are obtained from the ImageNet-22k dataset, namely the ImageNet-21k dataset plut the ImageNet-1k dataset.}
  \label{bsds_pre}
  \setlength{\tabcolsep}{2pt}
  \begin{tabular}{c c c c c}
    \toprule
    Methods & Pre-training & ODS & OIS & AP \\
    \midrule
    BDCN$^{paper}_{VGG}$ & ImageNet-1k & 0.828 & 0.844 & 0.890 \\
    \midrule
    BDCN$^{paper}_{ResNet}$ & ImageNet-1k & 0.832 & 0.847 & 0.872 \\
    \midrule
    BDCN$^{testing}_{VGG}$ & ImageNet-1k & 0.826 & 0.842 & 0.853 \\
\midrule
    EDTER$^{paper}$ & ImageNet-22k & 0.848 & 0.865 & 0.903 \\
    \midrule
    EDTER$^{re-training}$ & ImageNet-22k & 0.847 & 0.863 & 0.855\\
    \midrule
    msmsfnet$^{pre-training}$ & ImageNet-1k & 0.839 & 0.853 & 0.875 \\
    \midrule
    \midrule
    BDCN$^{random}_{VGG}$ & - & 0.803 & 0.821 & 0.837 \\
    \midrule
    EDTER$^{random}$ & - & 0.736 & 0.769 & 0.776 \\
     \midrule
     msmsfnet$^{random}$ & - & 0.823 & 0.840 & 0.858 \\
    \bottomrule
    \end{tabular}
\end{table}

The performance obtained by BDCN, EDTER, and the proposed msmsfnet in the BSDS500 dataset can be found in table ~\ref{bsds_pre}.
From table~\ref{bsds_pre} we can see that the proposed msmsfnet achieves superior performance than BDCN when the pre-trained weights are used to initialize the models' parameters.
Even though EDTER achieves the best performance on the BSDS500 dataset in all three criteria, the pre-trained weights which are used for parameter initialization are obtained by pre-training the backbone network on the ImageNet-22k dataset, namely the ImageNet-21k dataset plus the ImageNet-1k dataset.
The ImageNet-21k dataset contains around 14 million images and is much larger than the ImageNet-1k dataset, which contains around 1.28 million images.
Therefore, the comparison between the performance of EDTER and the proposed msmsfnet when using the pre-trained weights is not fair, as much more data has been used to train EDTER. 
When all models are trained from scratch, it is clearly shown in table~\ref{bsds_pre} that the proposed msmsfnet achieves much better performance than BDCN and EDTER.
In particular, the performance achieved by EDTER is far below than that of BDCN and msmsfnet, around 7 percent lower than BDCN and 9 percent lower than msmsfnet in terms of ODS F1-score.
This can be expected as it has been proven in ~\cite{vit} that the vision transformer requires sufficient amount of data to obtain better performance than CNNs. 
In the absence of sufficient amount of data for training, the CNN-based models are better choices as they can obtain better performance.

%% file: conclusions.tex
\section{Conclusions}
\label{conclusions}
In this paper we focus on the task of edge detection in both natural images and SAR images. 
Unlike previous works that use pre-trained ImageNet models, 
we study the performance achievable by existing methods when the pre-trained weights are not used. 
The motivation behind this work is two-fold: first, the use of pre-trained weights in existing methods limits the design space of network architecture for edge detection, as the proposed methods have to be built upon existing well-trained ImageNet models;
second, the strategy to pre-train a model using a large dataset and then fine-tune the model using a small dataset, which is commonly used in the field of computer vision, is not always possible, as explained and demonstrated in the experiments for edge detection in SAR images. 
The experimental settings in this work facilitate the design of new network architectures for edge detection, as pre-training a model on the ImageNet dataset is not practical for many researchers due to limited computation resources. 
Furthermore, we propose a new model for edge detection, which adopts a multi-branch architecture to enhance its multi-scale representation capability and utilizes spatial asymmetric convolutions so as to increase its accuracy by developing a very deep model. 
The proposed msmsfnet achieves superior performance than state-of-the-art methods when all models are trained from scratch.
When the pre-trained weights are used for parameter initialization, the performance of the proposed msmsfnet improves further and is also competitive on the BSDS500 dataset, which shows the efficiency of the proposed method.